\documentclass{article} 
\usepackage{iclr2024_conference,times}

\usepackage{amsmath,amsfonts,bm}

\def\eqref#1{equation~\ref{#1}}

\def\1{\bm{1}}

\DeclareMathAlphabet{\mathsfit}{\encodingdefault}{\sfdefault}{m}{sl}
\SetMathAlphabet{\mathsfit}{bold}{\encodingdefault}{\sfdefault}{bx}{n}

\usepackage{hyperref}
\usepackage{url}

\usepackage{wrapfig}
\usepackage{graphicx}
\usepackage{amsmath}
\usepackage{makecell}
\usepackage{algorithm}
\usepackage{multirow}
\usepackage{colortbl}
\usepackage{booktabs}
\usepackage{tabularx}
\usepackage{xcolor}

\usepackage{subfigure, cleveref}
\usepackage{amsmath,multirow}
\usepackage{booktabs}
\usepackage{algorithm}
\usepackage[noend]{algpseudocode}
\algrenewcommand{\algorithmiccomment}[1]{\hfill\texttt{//} #1}
\algrenewcommand{\algorithmicrequire}{\textbf{Input:}}
\algrenewcommand{\algorithmicensure}{\textbf{Output:}}

\title{MosaicThinker: On-Device Visual Spatial Reasoning for Embodied AI via Iterative Construction of Space Representation}

\author{
Haoming Wang\textsuperscript{*} \& Qiyao Xue\textsuperscript{*} \& Weichen Liu \& Wei Gao \\
Department of Electrical and Computer Engineering \\
University of Pittsburgh \\
Pittsburgh, PA 15261, USA \\
\texttt{\{haw200, qix63, weichenliu, weigao\}@pitt.edu} \\
\thanks{\textsuperscript{*}The first two authors contributed equally to this work.}
}

\iclrfinalcopy 
\begin{document}

\maketitle
\vspace{-0.05in}
\begin{abstract}
When embodied AI is expanding from traditional object detection and recognition to more advanced tasks of robot manipulation and actuation planning, visual spatial reasoning from the video inputs is necessary to perceive the spatial relationships of objects and guide device actions. However, existing visual language models (VLMs) have very weak capabilities in spatial reasoning due to the lack of knowledge about 3D spatial information, especially when the reasoning task involve complex spatial relations across multiple video frames. In this paper, we present a new inference-time computing technique for on-device embodied AI, namely \emph{MosaicThinker}, which enhances the on-device small VLM's spatial reasoning capabilities on difficult cross-frame reasoning tasks. Our basic idea is to integrate fragmented spatial information from multiple frames into a unified space representation of global semantic map, and further guide the VLM's spatial reasoning over the semantic map via a visual prompt. Experiment results show that our technique can greatly enhance the accuracy of cross-frame spatial reasoning on resource-constrained embodied AI devices, over reasoning tasks with diverse types and complexities.	
\end{abstract}

\section{Introduction}

\begin{wrapfigure}{r}{2.7in}
	\centering
	\subfigure[Traditional visual tasks of object detection and recognition]{
		\includegraphics[width=0.85\linewidth]{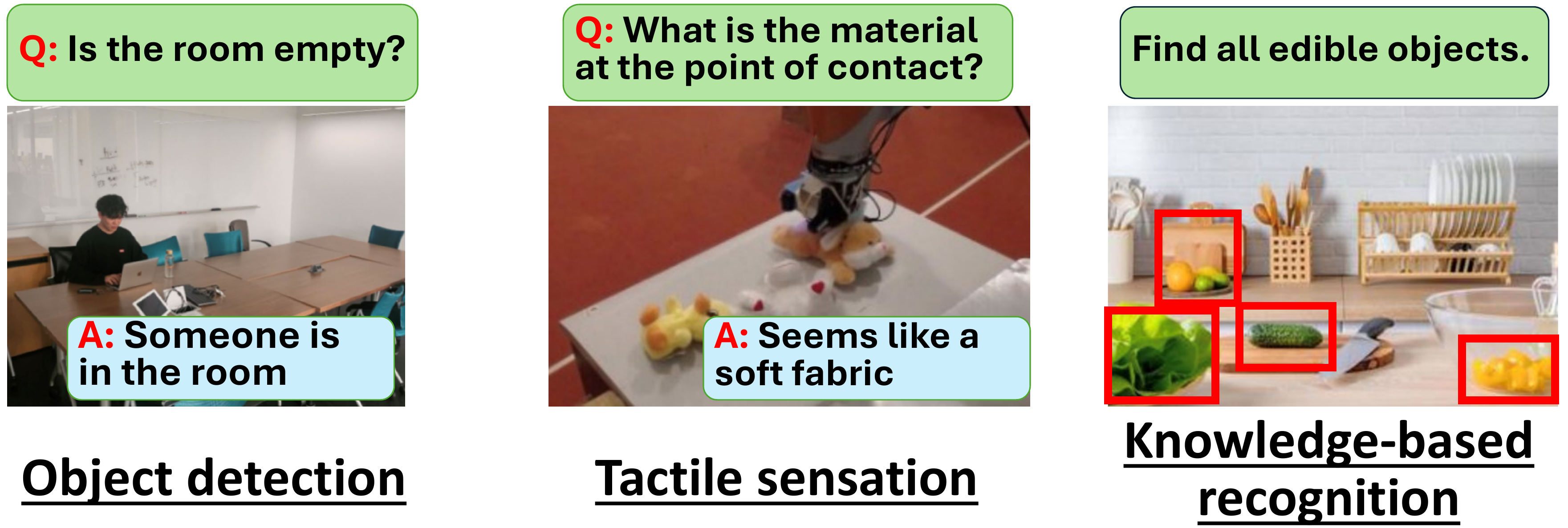}
		\label{fig:traditional_task}
	}
	\vspace{-0.1in}
	\subfigure[Visual spatial reasoning tasks]{
		\includegraphics[width=0.85\linewidth]{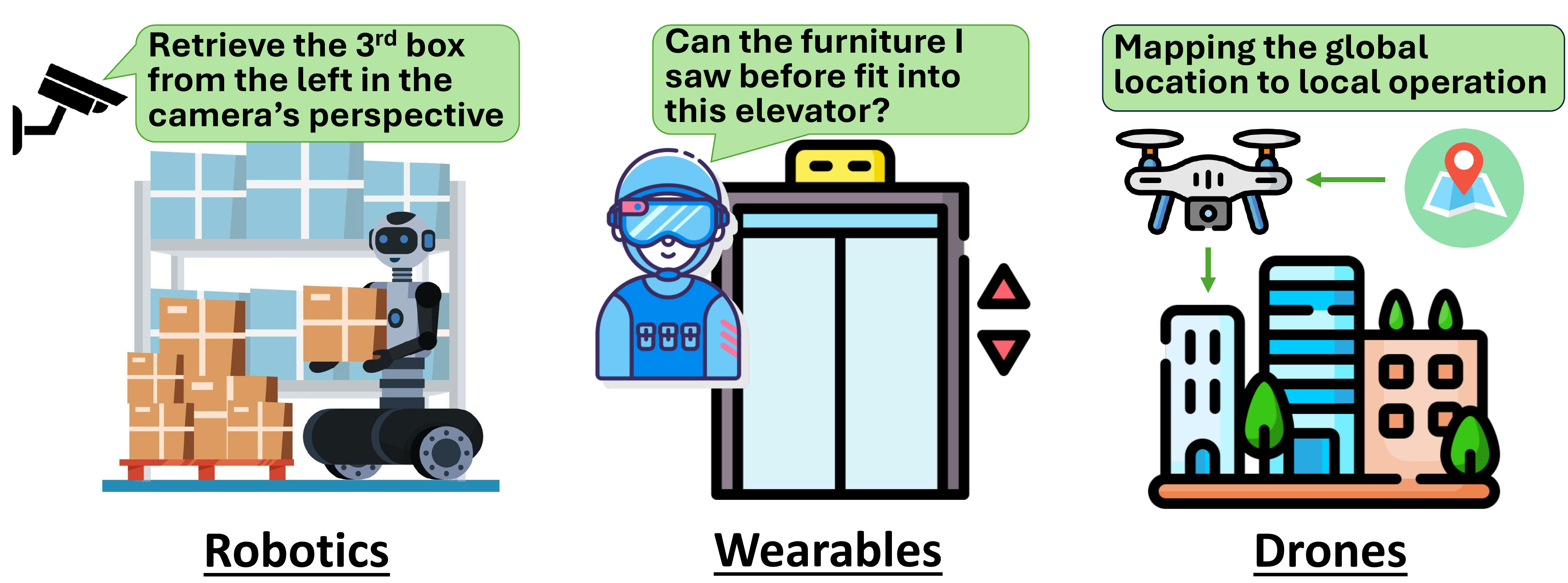}
		\label{fig:reasoning_task}
	}
	\caption{Different tasks in embodied AI applications}
	\vspace{-0.1in}
	\label{fig:all_scenarios}
\end{wrapfigure}

In embodied AI applications, a physical device captures visual views of the physical world into video streams, which are perceived by on-device AI models for proper device actions \cite{li2025challenges,wang2025empowering}. Compared to cloud-based solutions \cite{duan2022survey,driess2023palm,wang2024embodiedscan}, on-device AI models protects data privacy and ensures prompt device actions that are important in many mission-critical domains. Currently, the existing on-device AI models mostly focus on appearance-based object detection and recognition based on visual features such as texture, color and shape. For example in Figure \ref{fig:traditional_task}, semantic segmentation models can label every pixel to an object 
\cite{kirillov2023segment}, and detection models draw bounding boxes for objects \cite{liu2024grounding}. By further applying these visual inputs to an on-device Visual Language Model (VLM) \cite{hurst2024gpt,chu2024mobilevlm}, natural language queries can be answered by being grounded in visual perceptions \cite{pi2024perceptiongpt,yang2024llm}. These AI models have been applied to many embodied AI domains \cite{das2018embodied}, to drive tasks including object recognition in robotics \cite{zheng2025survey,levine2018learning} and visual assistance in augmented reality \cite{lin2023advances}.

Recently, the scope of embodied AI is further expanding to more advanced tasks such as robotic manipulation \cite{yuan2024cross,vuong2024language} and actuation planning \cite{zhang2025embodied,mao2024robomatrix}. In these tasks, VLMs also need to be capable of \emph{spatial reasoning} in the 3D space, such as comprehending the spatial relationships (e.g., orientation and relative position) of objects, to correctly interpret the spatial contexts and guide device actions \cite{jia2025omnispatial,ma2024survey,ogezi2025spare}. For example, in Figure \ref{fig:reasoning_task}, to identify the correct box to retrieve, a robot translates the allocentric input command into egocentric actions, requiring correct understanding of the spatial relationships across different viewpoints (camera's view and robot's view). Similarly, a wearable assistant estimates the object's 3D dimensions to determine whether it fits into the elevator, and a search-and-rescue drone builds a 3D map of an unknown area to report the survivor's location relative to the surrounding landmarks. Unfortunately, the existing VLMs perform very poorly in spatial reasoning tasks \cite{fu2024blink,wang2024muirbench}, especially from RGB video captures where spatial information is only contained in an implicit manner\footnote{RGB video captures are the only usable visual modality on many low-cost embodied AI devices with limited local resources. See \cref{subsec:other_modality} for details.}.

The existing VLMs fail in spatial reasoning due to the following reasons. First, current VLMs are mostly designed and trained to take 2D images as inputs, and inherently lack knowledge about 3D spatial information. These models encode visual inputs as flat feature maps or token sequences, without perceiving the object-centric 3D dimensions and spatial relationships. Hence, they can only capture simple spatial relations based on pixel coordinates (e.g., the relative position in camera view), but struggle with more complex relations that involve viewpoint transformations across multiple frames and require good alignment over frames.
Such cross-frame alignment is challenging because spatial information across different views presents as multiple egocentric observations, which are partial and fragmented \cite{you2024multiview,zhang2024multiview} (\cref{subsec:cross_frame}). We verified that VLM's ability of spatial reasoning scales with parameter size (\cref{subsec:spatial_reasoning_vlm}), and such ability of small on-device VLMs is even weaker. 

	

An intuitive solution to these challenges is to redesign a VLM for spatial reasoning tasks, and train the VLM with 3D spatial data. However, model redesign and training are expensive, and this data-driven approach has poor generalizability in different task domains that differ in scene layouts and instruction distributions \cite{sun2024survey,zitkovich2023rt}.

\begin{wrapfigure}{r}{2.7in}
	\centering
		\vspace{-0.1in}
	\includegraphics[width=\linewidth]{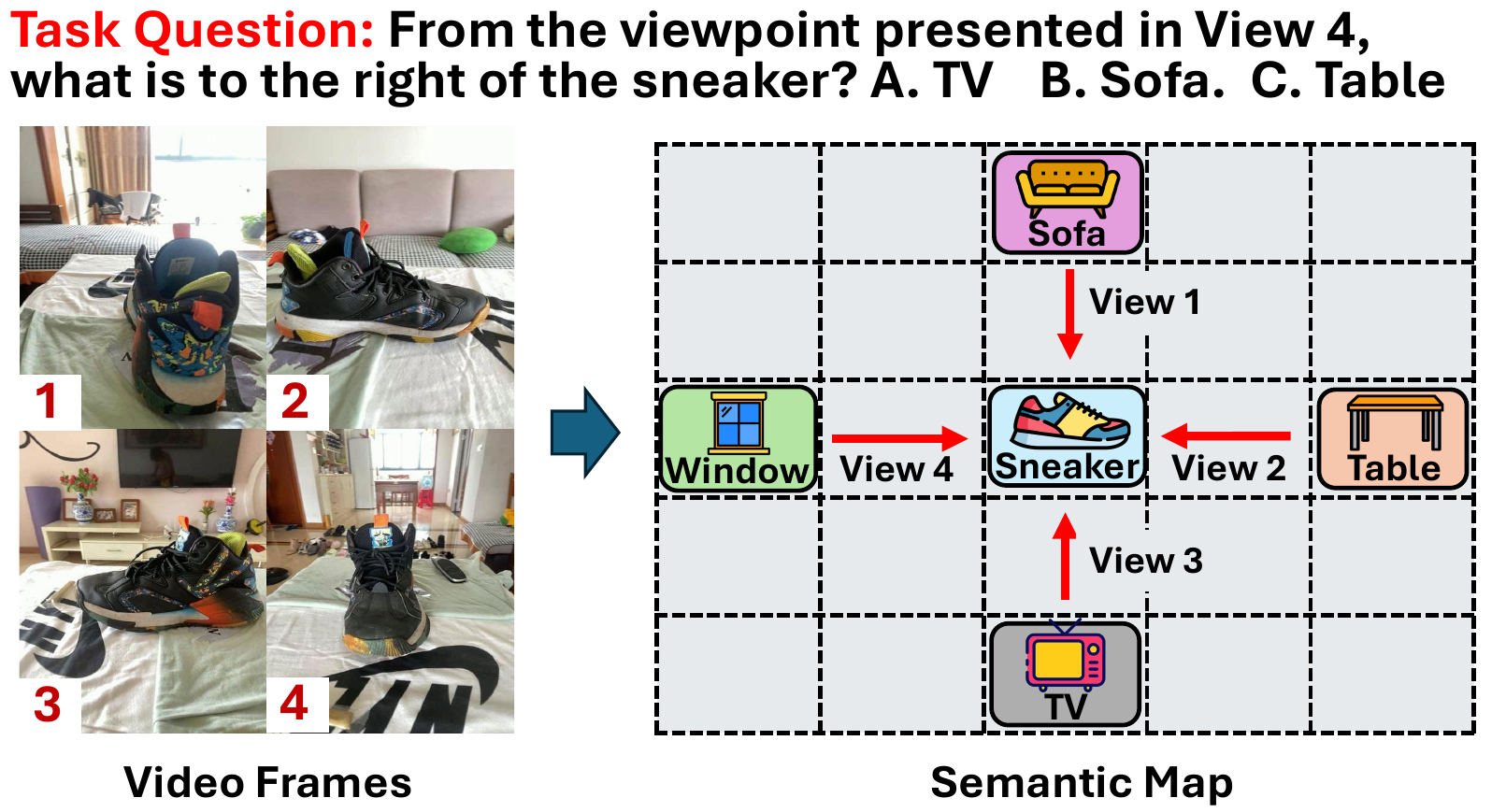}
	\vspace{-0.3in}
	\caption{Construction of semantic map as the unified space representation}
	\label{fig:semantic_map_eg}
\end{wrapfigure}

Rather than retraining VLMs, in this paper we present a new inference-time computing technique, namely \emph{MosaicThinker}, which properly structures the available spatial information and accordingly guides the VLM's process of spatial reasoning. To ensure good cross-frame alignment, our key idea in MosaicThinker is to integrate fragmented spatial information from multiple frames into a unified space representation. Being different from traditional methods of dense representations such as bird's-eye view (BEV) maps \cite{qi2025gpt4scene} that are hard for small on-device VLMs to perceive, in MosaicThinker we construct a sparse representation as a \textbf{global semantic map}. As shown in Figure \ref{fig:semantic_map_eg}, the semantic map represents rich spatial information about locations of camera and all the task-related objects across different frames as a sparse grid, rather than an image of dense top-down pixels. As a result, we can enable effective cross-frame spatial reasoning by instructing the small on-device VLMs to perceive the semantic map via a carefully crafted visual prompt (\cref{subsec:visual_prompt}). For example in Figure \ref{fig:semantic_map_eg}, 
the task-related object (i.e., the object right to the sneaker) is missing in the current view (view 4), but using the semantic map allows easy VLM reasoning, 
as the VLM can reference the object's global position on the map relative to the camera's current pose.

For small on-device VLMs with limited representation power, it is hard to construct the semantic map by directly prompting the VLM to retrieve the spatial information from frames and establish cross-frame spatial correlations. Instead, we first use specialized AI models to extract spatial information from individual frames, and then iteratively aligns such information across consecutive frames by matching objects in different views (\cref{subsec:iterative_construction}). In this way, this alignment fuses the per-frame spatial information into a unified global coordinate system, from which the semantic map can be coherently constructed with all the objects' locations (\cref{subsec:iterative_construction}).

Intuitively, all frames in the input video could be used for constructing the semantic map, but doing so is not only computationally expensive but also semantically noisy, as the semantic map may improperly contain irrelevant objects that affect reasoning on task-related objects. Instead, we only involve a set of key frames relevant to the reasoning task in construction of semantic map. To efficiently identify the most proper set of key frames, MosaicThinker samples the video frames in multiple iterations, and progressively refines its sampling distribution in each iteration to focus on the most informative video segments (\cref{subsec:keyframe_selection}).

To our best knowledge, MosaicThinker is the first training-free method that enables complicated cross-frame visual spatial reasoning on small on-device VLMs. 
We implemented MosaicThinker on on-device AI platforms, including NVidia Jetson Orion, Meta AR Glass and OnePlus 12R smartphone. We evaluated MosaicThinker with multiple spatial reasoning benchmarks that cover different indoor scenes (e.g., residential homes, offices, libraries, etc), focusing on various types of spatial reasoning tasks about object relationship, location identification and camera movement estimation. From our experiment results, we have the following conclusions:
\begin{itemize}
	\item MosaicThinker significantly improves the accuracy of difficult cross-frame spatial reasoning tasks. Compared to competitive baselines of video understanding and spatial reasoning, it can improve the reasoning accuracy by up to 40\% on various types of tasks. 
    \vspace{0.05in}
	\item MosaicThinker is highly adaptive. The proposed method of semantic map construction can well adapt to different types of indoor scenes and object complexity with possible occlusion and partial visibility. 
    \vspace{0.05in}
	\item MosaicThinker is lightweight. When deployed with off-the-shelf VLMs on embodied AI devices, it retains high compute efficiency and incurs a small amount of extra compute overhead.
\end{itemize}

\vspace{-0.1in}
\section{Background and Motivation}


\subsection{Spatial Reasoning on Small VLMs}
\label{subsec:spatial_reasoning_vlm}

\begin{wrapfigure}{r}{3in}
	\centering
	\vspace{-0.1in}
	\includegraphics[width=\linewidth]{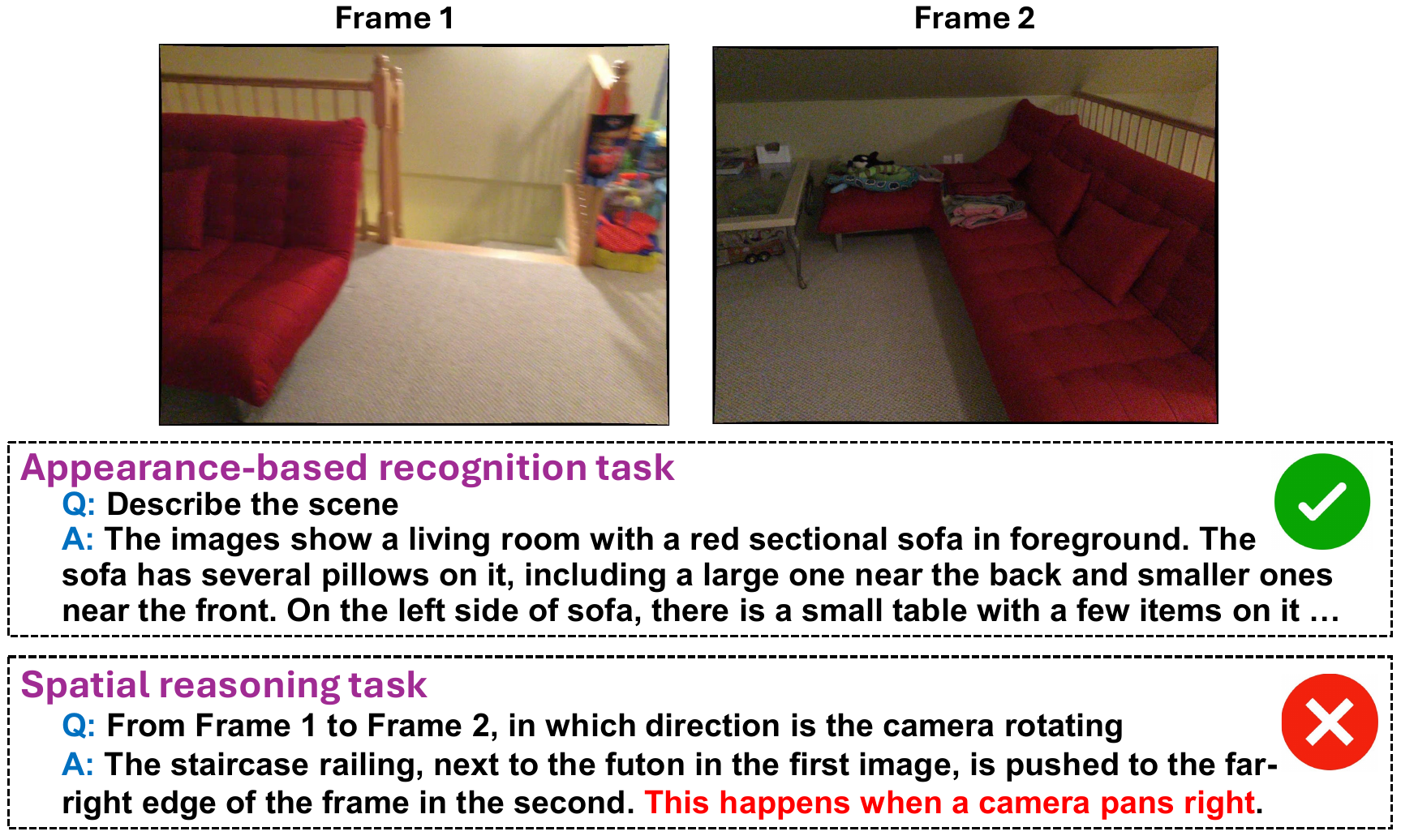}
	\vspace{-0.25in}
	\caption{The existing VLM's performance discrepancy between appearance-based and spatial reasoning tasks. Outputs are generated by Qwen-2.5-VL-32B.}
	\label{fig:vlm_spatial_reasoning}
	\vspace{-0.15in}
\end{wrapfigure}

VLMs exhibit near-human performance in appearance-based tasks such as semantic understanding and open-vocabulary recognition, but they struggle with spatial reasoning tasks that are relatively easy for humans \cite{yang2025mmsi,yang2025thinking}, as exemplified in Figure \ref{fig:vlm_spatial_reasoning}. This discrepancy arises from the fact that current VLMs learn primarily from static texts and 2D images, which rarely contain explicit information about the 3D space that humans continuously perceive. Moreover, cognitive science has verified that the human brain perceives the space by converting the egocentric 3D perceptions into an allowcentric \emph{mental map}, and hence facilitates reasoning from multiple viewpoints \cite{johnson1980mental,johnson1983mental}. In contrast, VLMs process texts and visual features in sequential tokens without an explicit representation of 3D space, limiting their ability of spatial reasoning. We are hence inspired to construct a semantic map as a 3D space representation that encodes objects' positions, orientations and spatial relationships, to mimic the human brain's reasoning on the mental map.

\begin{wrapfigure}{r}{3in}
    \centering
    \includegraphics[width=0.95\linewidth]{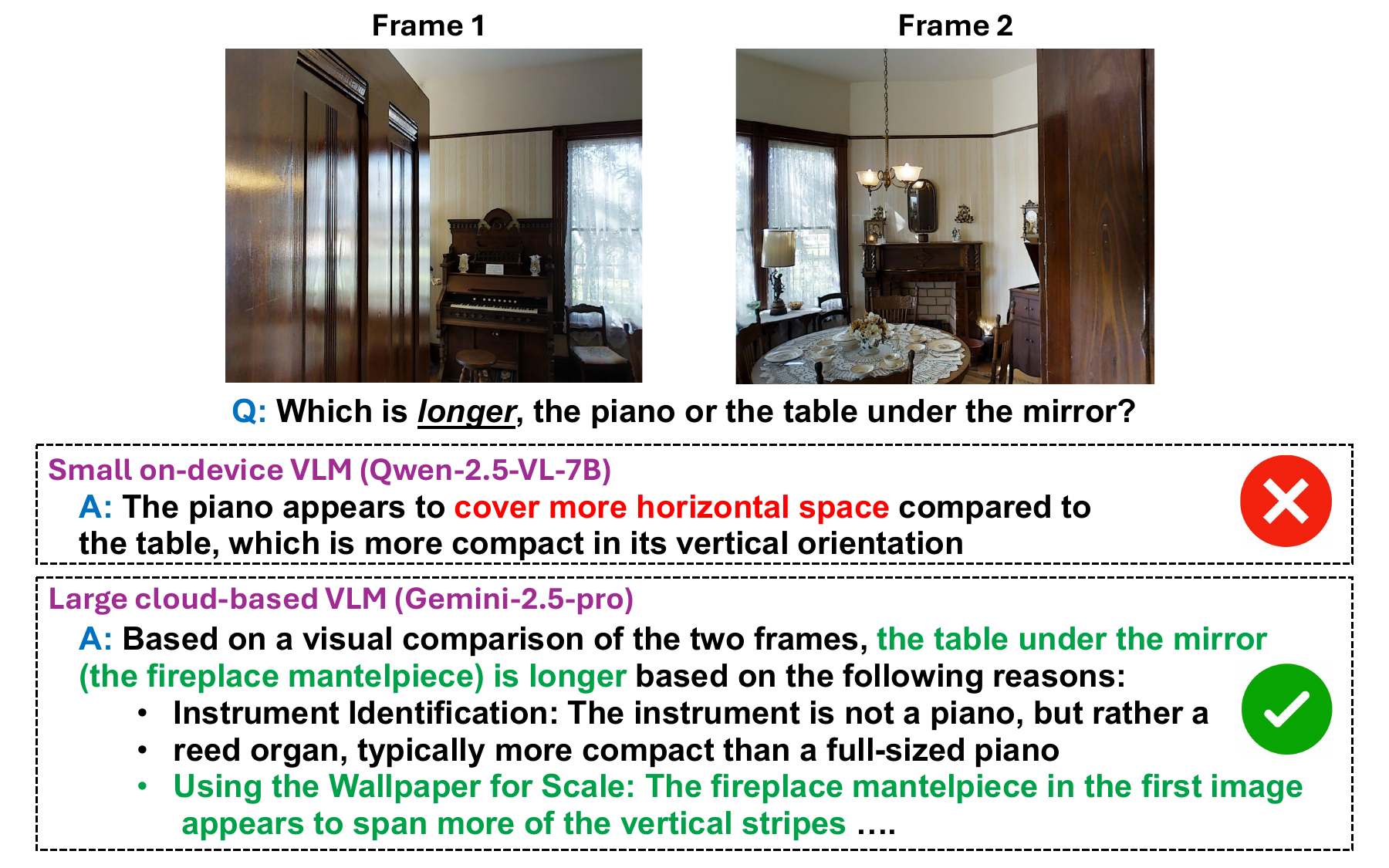}
    \vspace{-0.1in}
    \caption{The VLM's spatial reasoning capability reduces with the model size}
    \label{fig:small_large_vlm}
    \vspace{-0.2in}
\end{wrapfigure}

In addition, the spatial reasoning ability of VLMs also scales with the model size \cite{yang2025mmsi,yang2025thinking}. Existing studies show that, on relatively simple tasks such as estimating the object sizes, a state-of-the-art VLM (e.g., Gemini-2.5-pro) can achieve human-level performance, but a small VLM with 3B parameters barely outperforms a random guess \cite{fu2024blink,wang2024muirbench}. As shown in Figure \ref{fig:small_large_vlm}, larger VLMs usually encode visuospatial commonsense knowledge such as the typical size of a chair, and can hence reason about the sizes of other objects by comparing their relative scale. Smaller VLMs, in contrast, contain only limited knowledge about such commonsense. Clearly, deploying a large VLM on embodied AI devices is impractical, and this constraint motivates us to improve the 
spatial reasoning capability of small on-device VLMs beyond their representation power.

\vspace{0.1in}
\subsection{Cross-frame Spatial Reasoning}
\label{subsec:cross_frame}
\begin{wrapfigure}{r}{3in}
\centering
	\vspace{-0.1in}
\includegraphics[width=\linewidth]{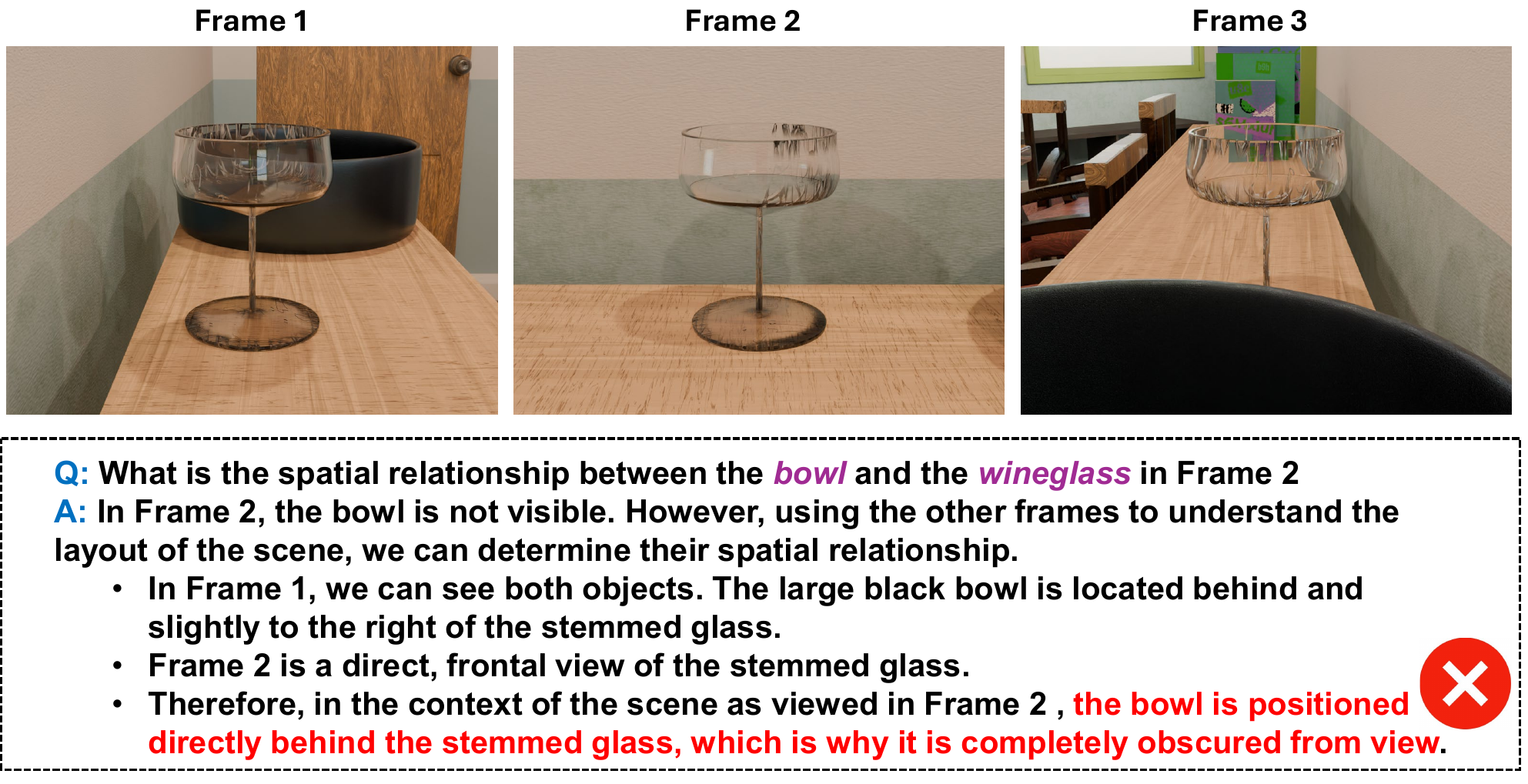}
\vspace{-0.25in}
\caption{Incorrect cross-frame spatial reasoning where task-related objects are occluded or separately appear in frames. Outputs are generated by Gemini-2.5-Pro.}
\label{fig:occluded_view}
\vspace{-0.1in}
\end{wrapfigure}

One common approach to enhancing the VLM's capability of spatial reasoning is to extract 3D spatial information using a 3D foundation model and inject it into the VLM, either explicitly through a depth map or implicitly via high-level features \cite{wu2025spatial,huang2025mllms,zheng2025learning,wang2025ross3d}. Some existing methods proposed more fine-grained injection, e.g., segmenting task-related objects and encoding them as separate spatial tokens \cite{yu2025inst3d}, or dividing the scene into regions and focusing on those most relevant to the task \cite{zhi2025lscenellm}. However, these methods still struggle with complicated reasoning tasks across multiple frames, because the task-related objects may separately appear in different frames or be partially occluded across frames, as shown in Figure \ref{fig:occluded_view}. In these cases, partial observations across frames cannot be well integrated into a coherent space representation, and the reasoning process hence remains fragmented and error-prone. Tackling this type of challenging spatial reasoning tasks will be the main focus of this paper.

Recently, some research efforts aim to address this challenge by unifying the space representations from multiple frames. For example, generating a bird's-eye view (BEV) map has been effective in consistently marking objects across frames, when being applied to large VLMs \cite{qi2025gpt4scene}. However, this approach performs poorly for small on-device VLMs \cite{qi2025gpt4scene}, because the BEV map is a dense representation of the 3D space, which is hard for small VLMs with limited representation power to interpret. Instead, some existing work suggests to use Scene Graph Reconstructions as the sparse representation \cite{gao2022classification,yang2023panoptic}, which however, requires dense supervision, closed-set labels and task-specific architectures, hence lacking generalizability. These limitations, then, motivates us to design semantic-based methods of space representation, for better sparsity and generalizability.


	
	

	\vspace{-0.05in}
	\subsection{Other Visual Modalities}
	\label{subsec:other_modality}
	Another approach to enhancing cross-frame spatial reasoning is to incorporate additional visual modalities, such as depth map \cite{zhu2024llava} and point clouds \cite{linghu2024multi}. Dedicated hardware can provide these visual inputs, but devices with sufficient accuracy are often too expensive for low-cost embodied AI devices such as small robots and AR/VR headsets \cite{simon2023mononav,skurowski2024energy}. Specialized AI models can also estimate these visual modalities from RGB captures \cite{yang2024depth,bochkovskii2024depth}, but their estimation accuracy substantially degrades in the far field
	\cite{jiao2018look,zhan2025vistadepth}. For instance, experiments in \cite{zhan2025vistadepth} show that errors of depth estimation largely rises once the normalized depth range exceeds 60\%. Due to this discrepancy, rather than directly using the raw depth estimations to calculate object locations in the semantic map, in MosaicThinker we iteratively align depth estimates (from RGB video captures) across frames, and only use the aligned depth point to construct the semantic map.

	Furthermore, even when accurate depth maps or point clouds are available, they cannot be directly used as visual input to small on-device VLMs, which are not trained to interpret these modalities. For example, if we apply the depth map to VLMs as a RGB image, small VLMs with 3B-7B parameters can only extract coarse-grained depth information for categorizing the object's position (e.g., foreground, midground or background) \cite{mukhopadhyay2024mapwise,rahmanzadehgervi2024vision}. This deficiency is another motivation for us to instead construct a sparse spatial representation of semantic map, to better guide VLM's reasoning through carefully designed visual prompts.

	\begin{figure*}
    \includegraphics[width=0.9\textwidth]{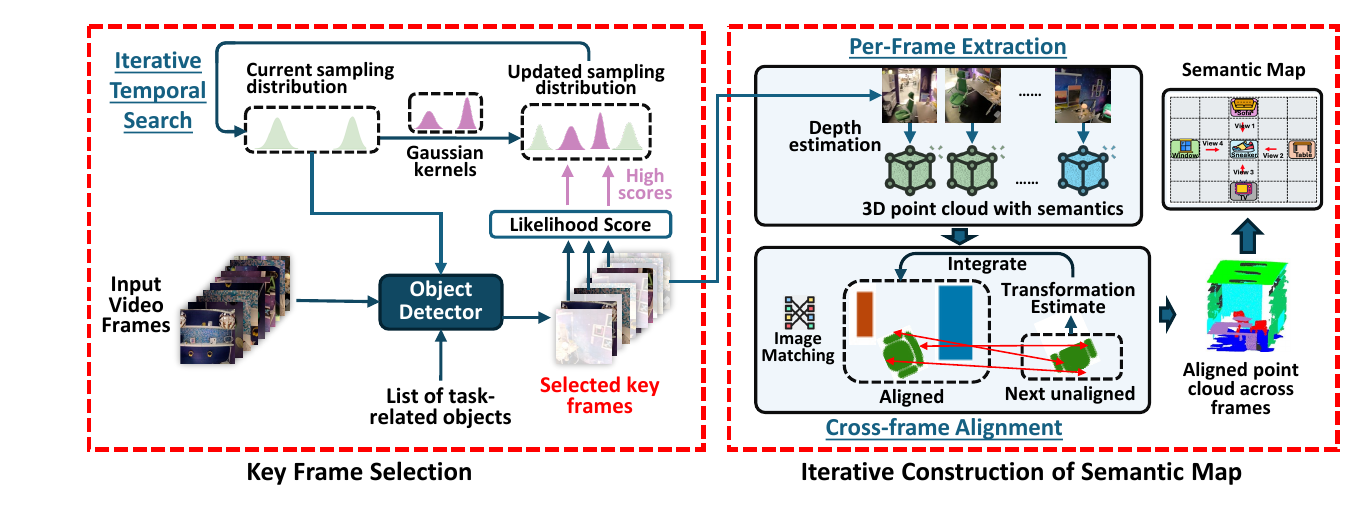}
    \vspace{-0.1in}
    \caption{Overview of MosaicThinker design}
    \label{fig:overview}	\vspace{-0.2in}
\end{figure*}

\section{MosaicThinker Design}
	\label{sec:overview}
	
	MosaicThinker targets spatial reasoning tasks that takes an egocentric video of the scene as visual input and answers a user's question in natural language, as exemplified in Figure \ref{fig:vlm_spatial_reasoning}, and consider a ``cold start'' scenario, where the VLM has no prior knowledge in any form about the scene to be reasoned. MosaicThinker focuses on cross-frame reasoning of \emph{object-centric} spatial relationships (e.g., left/right/front/behind, containing, supporting, adjacent to, etc), rather than measurements of dimensions and distances \cite{chen2024spatialvlm, jia2025omnispatial, ogezi2025spare}. We consider indoor scenes with moderate camera motions and occasional occlusions in video captures, as rapid camera motions cause motion blurs in videos and result in corrupted video frames, and frequent occlusions could significantly reduce the amount of available spatial information about task-related objects in the scene, making the spatial reasoning practically infeasible. 

	As shown in Figure~\ref{fig:overview}, MosaicThinker consists of two major components: (1) \emph{Iterative Construction of Semantic Map}, which incrementally builds a unified 3D space representation that contains rich spatial information about task-related objects; and (2) \emph{Key Frame Selection}, which identify a minimum yet sufficient set of video frames as visual inputs for constructing the semantic map. MosaicThinker then uses the semantic map as input to on-device VLM, and uses a carefully crafted visual prompt to instruct the VLM's reasoning over the semantic map.
	

\subsection{Preprocessing} 

\begin{wrapfigure}{r}{2.8in}
	\centering
	\vspace{-0.3in}
	\includegraphics[width=\linewidth]{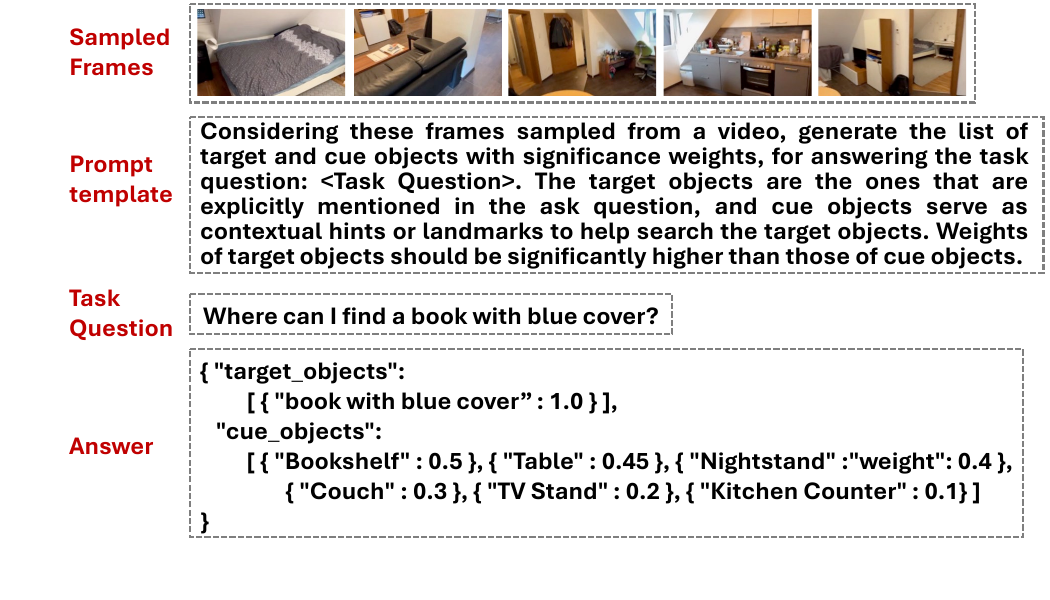}
	\vspace{-0.25in}
	\caption{An example of task grounding generated by Gemini-2.5-pro}
	\label{fig:target_grounding}
	\vspace{-0.1in}
\end{wrapfigure}

Given a reasoning task, MosaicThinker first obtains the list of task-related objects in natural language, by grounding the task question using a VLM. As shown in Figure \ref{fig:target_grounding}, instead of text parsing or applying the entire video to VLM, we prompt the VLM with the task question and a few randomly sampled video frames, to disambiguate object aliases and scene context. The VLM is then instructed to output a list of object names in natural language, ranked by their relevance to the question and likelihoods of appearance in the video, based on joint understanding of question semantics and visual priors inferred from the sampled frames.

This list is expected to include 1) \emph{target objects} that are explicitly mentioned in the question (e.g., ``book with blue cover''), and 2) \emph{cue objects} that serve as contextual hints or landmarks to facilitate spatial reasoning (e.g., ``bookshelf''). Since this grounding relies solely on appearance-based recognition, an on-device VLM is sufficient.

Besides, for better efficiency of reasoning, we also preprocess the input video frames to detect and remove redundant frames that are identical or highly similar. Such redundancy can be calculated as the pairwise similarity between frames using various metrics such as pixel-wise difference \cite{zhu2013video} or histogram correlation \cite{zweng2011evaluation}, but is highly expensive if applied to all frames. To improve the compute efficiency, we adopt lightweight filtering techniques, by first applying uniform down-sampling and motion-based keypoint tracking with ORB/SIFT \cite{rublee2011orb} to sequentially identify candidate changes, and then performing more costly similarity checks (e.g., SSIM \cite{wang2004image}) only on this subset. 

\subsection{Iterative Construction of Semantic Map}
\label{subsec:iterative_construction}

\begin{wrapfigure}{r}{3in}
	\centering
		\vspace{-0.2in}
	\includegraphics[width=\linewidth]{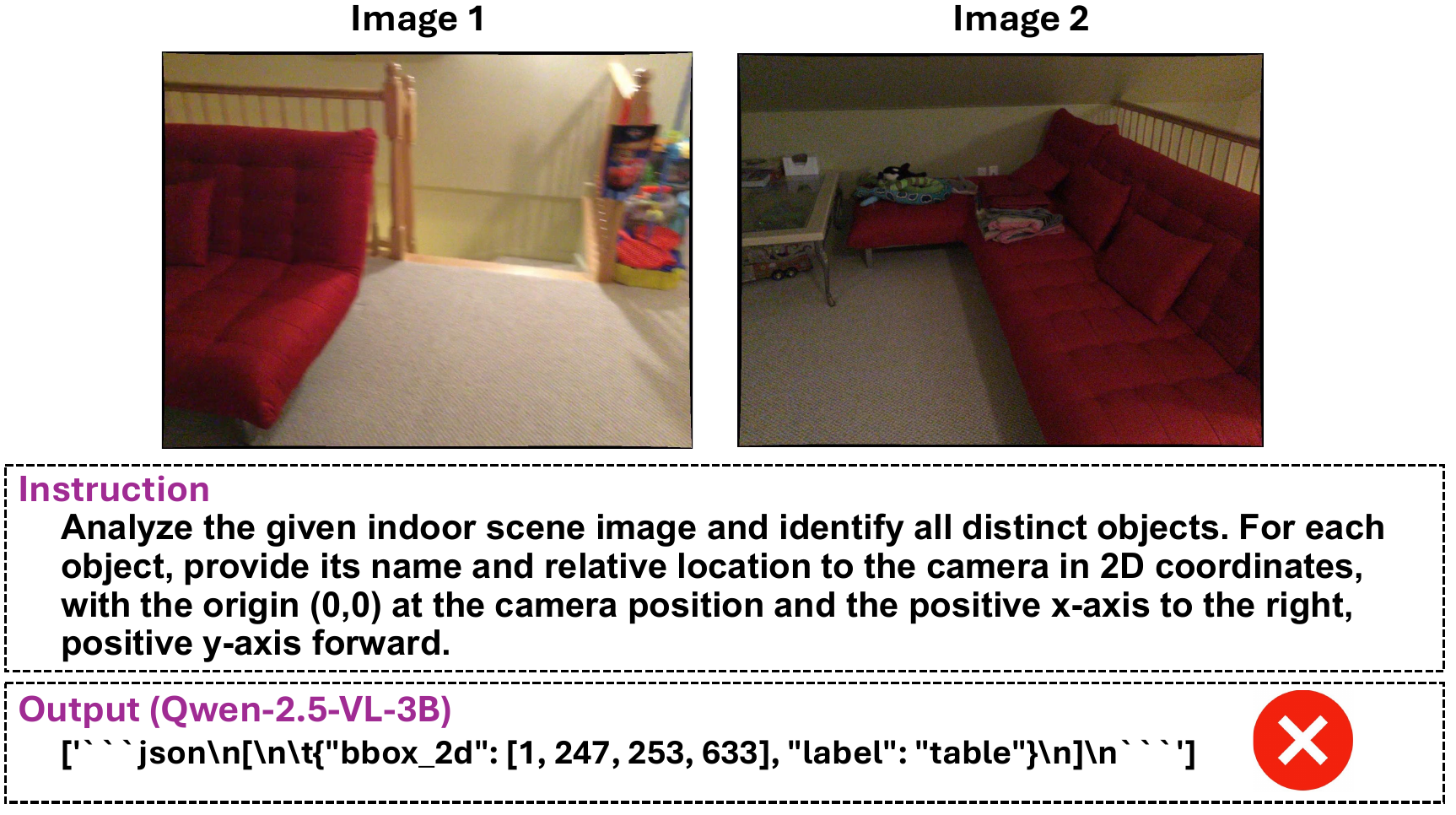}
	\vspace{-0.2in}
	\caption{Ineffective generation of the semantic map by directly prompting the small on-device VLM}
	\label{fig:semantic_map_direct_gen}
	\vspace{-0.1in}
\end{wrapfigure}

Construction of the semantic map is to decide the locations and orientations of camera and all objects in the scene, and map them into a global sparse grid. One method is to directly prompt the VLM to output such information given the video frames, in a structured format. However, small on-device VLMs lack sufficient representation power to accurately infer 3D measurements and relative positions from 2D images. As shown in Figure \ref{fig:semantic_map_direct_gen}, the list of objects generated by the VLM is incomplete, and the reported locations are clearly wrong.

Another alternative is to explicitly extract such spatial information using specialized AI models. For each frame, we can first use a segmentation model to identify individual objects, and then use a depth estimation model to construct a 3D point cloud of the scene, from which the locations and orientations of objects in the frame can be computed. In practice, many existing open-sourced models can be used for such extraction with sufficient accuracy. For example, a lightweight segmentation model like MobileSAM with 1B parameters can achieve 75\% mIoU (Intersection over Union) on the COCO dataset \cite{zhang2023mobilesamv2}, and a small depth estimation model like ZoeDepth-N with 3.4B parameters can achieve a relative error of 0.05 on the NYU Depth V2 benchmark \cite{bhat2023zoedepth}.


Constructing the semantic map using the extracted spatial information, however, is still challenging, because the location and orientation of the same object in different frames may be different (e.g., the red sofa shown in Figure \ref{fig:semantic_map_direct_gen}). \emph{Cross-frame alignment} is hence needed to coherently place all objects into a global coordinate system. Intuitively, this global coordinate system could be set as the local coordinate system of the video frame that is most likely to contain the target object, to ensure that the constructed semantic map is centered at the target object, as shown in Figure \ref{sec:overview}.


To align visual information across frames, the first step is to estimate the camera transformation matrix $\mathbf{T}_{i \to j} \in SE(3)$ between a pair of frames $i$ and $j$, where $SE(3)$ denotes the space of 3D rigid transformations \cite{spong2006robot} containing translation and rotation information. Specifically, the transformation can be decomposed as:
\vspace{-0.05in}
\[
\mathbf{T}_{i \to j} =
\begin{bmatrix}
\mathbf{R}_{i \to j} & \mathbf{i}_{t \to j} \\
\mathbf{0}^\top & 1
\end{bmatrix},
\]

where $\mathbf{R}_{i \to j} \in SO(3)$ is the rotation matrix and $\mathbf{t}_{i \to j} \in \mathbb{R}^3$ is the camera translation between frames. Then, by applying these transformations, points observed in the local coordinate system of any frame can be projected into a unified global coordinate system, thereby fusing fragmented views into a coherent cognitive map.



The transformation matrix $\mathbf{T}_{i \to j} \in SE(3)$ could be computed using traditional iterative closest point (ICP) algorithms for point cloud matching  \cite{dai2016bundlefusion}, but these methods would be inaccurate when being applied to on-device deployment with limited compute power, which only allows a limited number of video frames for segmentation and depth estimation. In these cases, overlapping regions between these frames are often small, resulting in inaccurate matching. Instead, we perform matching directly to RGB images using a pretrained image matching model such as MatchAnything \cite{he2025matchanything}. Since the same object observed from different viewpoints usually retain consistent color patterns, a simple pixel-based similarity metric would be sufficient, and the low compute costs of image matching hence allows us to perform matching over all the given video frames. 
More details about such matching-based computation of the transformation matrix are presented in Section \ref{subsec:cross_frame_alignment_details}.

To integrate pairwise estimates of frame alignment into a unified global coordinate system, we replace naive sequential integration, which is susceptible to cumulative drift and fails when temporally adjacent frames lack visual similarity, with a topology-aware alignment strategy. More specifically, we organize video frames into a tree structure rooted at a central ``global anchor'' frame, and edges between nodes in the tree represent the maximum visual overlap between frames. By computing the global poses via the unique paths to the root rather than the frames' chronological order in the video, we ensure that transformations are derived exclusively from high-confidence pairs, effectively preventing error propagation and isolating outliers to the branches' ends. More details of such multi-frame alignment are in Section \ref{subsec:multi_frame_alignment}.

Another challenge that may affect the matching accuracy is the possible occlusion or partial visibility of objects, which often prevents the segmentation model from detecting them. 
To address this challenge, after having aligned all the frames, we estimate the global pose (including location and orientation) of the camera for each frame, based on the calculated transformation matrix. Using the camera parameters, we can infer which objects should fall within the current field of view (FoV) but were missed by segmentation. The current matching results are then used to guide the segmentation model to recover these occluded or partially visible objects. More details are provided in Section \ref{subsec:final_refinement}.

	\subsection{Key Frame Selection}
	\label{subsec:keyframe_selection}

	Intuitively, the semantic map can be constructed with all video frames. However, doing so is not only expensive but also semantically noisy, because long stretches of irrelevant frames can overweight unimportant task-irrelevant information (e.g., scene backgrounds and irrelevant objects) and affect reasoning on task-related objects. Instead, we construct the semantic map using only a set of key frames relevant to the reasoning task. 
	These key frames should contain as many task-related objects as possible, and each frame's likelihood of containing these objects is independently calculated as the weighted sum of detection confidence scores over these objects, calculated by a pre-trained object detector such as GroundingDINO \cite{liu2024grounding} and Yolo-World \cite{cheng2024yolo}. 
	An intuitive approach to key frame selection, then, is to calculate the likelihood scores of all frames and select the frames with the highest scores. However, this approach needs to score every frame and is computationally expensive. 
	

Instead, we select key frames via an iterative process of temporal search. As shown in Figure \ref{fig:overview}, in each iteration, we only score a small set of frames following a sampling distribution, and select them as key frames if their scores are higher than a pre-defined threshold. This sampling distribution is initialized as uniform, and then iteratively refined based on the temporal locality of object appearance among frames. That is, if a frame has a high likelihood score, other frames that are temporally close could also possibly have high likelihood scores, due to the continuity of camera movement in the video \cite{manasyan2025temporally, zhang2025continuous, wu2025video}. More specifically, for every frame that has been scored, we apply a Gaussian kernel to this frame's index in the sampling distribution, with the Gaussian kernel's amplitude and standard deviation decided by the similarity between frames, to boost up the chances of other frames before or after this frame to be sampled and scored. More details of such sampling distribution are in Section \ref{subsec:gaussian_kernel}.

In practice, we pre-define the number of key frames, based on the available on-device compute power and the required latency of reasoning. The efficiency of key frame selection also depends on the number of iterations and the threshold of likelihood score used in each iteration. These details of key frame selection are in Section \ref{subsec:selection_design_choice}.




	
\begin{wrapfigure}{r}{3.5in}
	\centering
	\includegraphics[width=\linewidth]{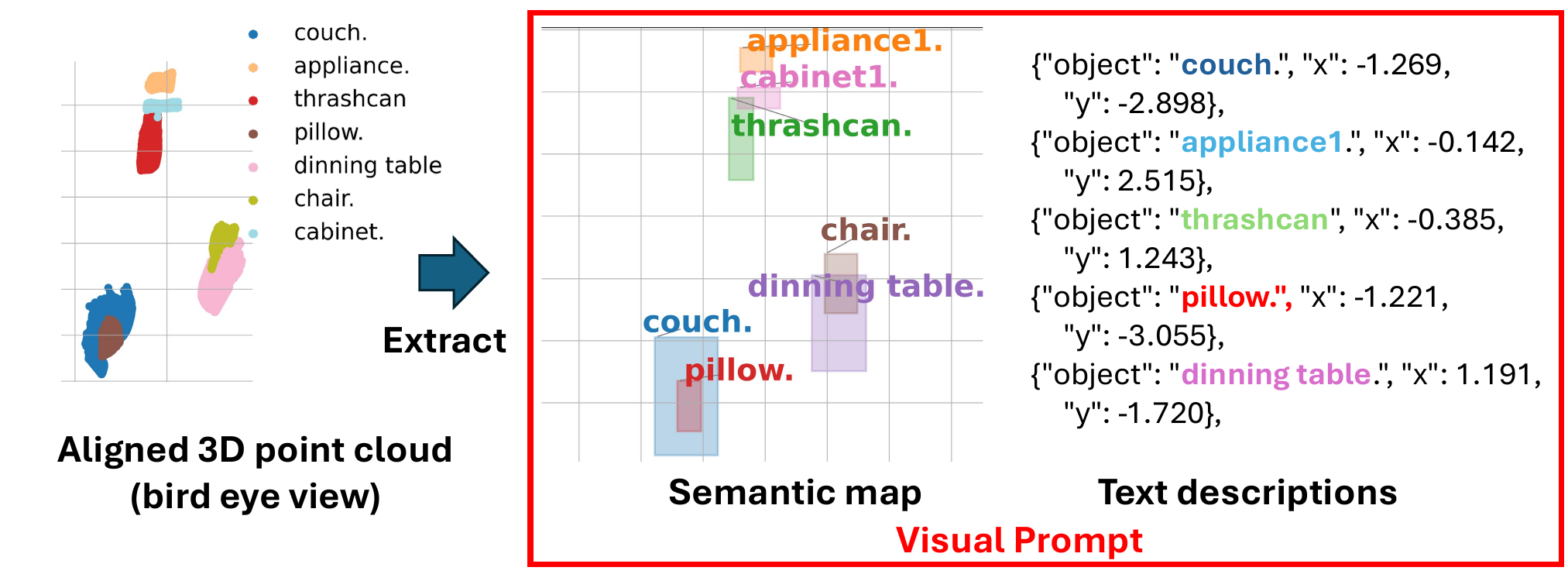}
	\vspace{-0.25in}
	\caption{Crafting the visual prompt from the aligned 3D point cloud}
	\label{fig:visual_prompt}
	\vspace{-0.1in}
\end{wrapfigure}

\subsection{Spatial Reasoning via Visual Prompts}
\label{subsec:visual_prompt}
The constructed semantic map is provided as input to the VLM, which is guided by a carefully crafted visual prompt to support spatial reasoning. As shown in Figure \ref{fig:visual_prompt}, the visual prompt represents the semantic map as a top-down grid with simple symbols illustrating the relative locations of objects from a BEV view, and is also accompanied by textual descriptions about the absolute location of each task-related object's bounding box\footnote{Here, we use bounding boxes rather than contours to eliminate the ambiguity in the BEV semantic map.} in the semantic map. These bounding boxes are computed from the point cloud with semantics as extracted in Section \ref{subsec:iterative_construction}, and each is assigned with a unique color for the VLM to distinguish between objects.

\begin{wrapfigure}{r}{3in}
	\centering
	\vspace{-0.25in}
	\includegraphics[width=1\linewidth]{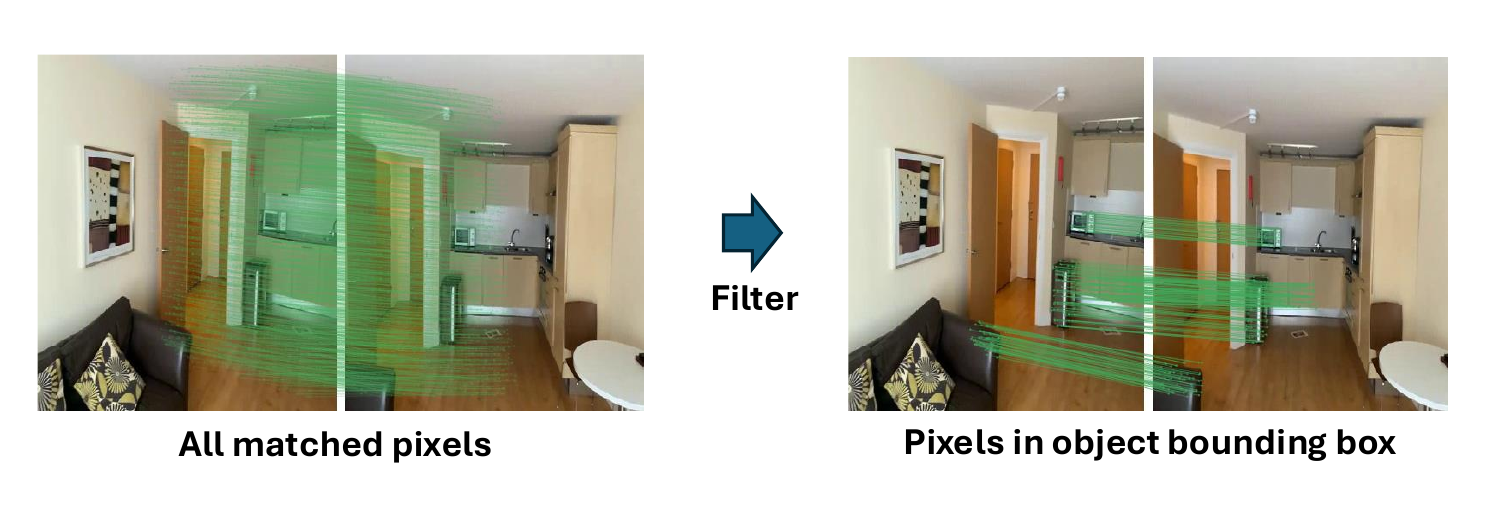}
	\vspace{-0.35in}
	\caption{Pixels pairs found by the image matching model}
	\label{fig:matched_pixels}
	\vspace{-0.1in}
\end{wrapfigure}

Note that, the semantic map can also be provided to the VLM in other ways, including plain text descriptions about locations of objects and camera, or dense point clouds rendered from a BEV view. However, these methods are usually too ambiguous for the small on-device VLM to correctly interpret. More details about these different design choices are discussed in Appendix \ref{subsec:visual_prompt_details}.

\begin{wrapfigure}{r}{3in}
	\centering
	\vspace{-0.3in}
	\includegraphics[width=\linewidth]{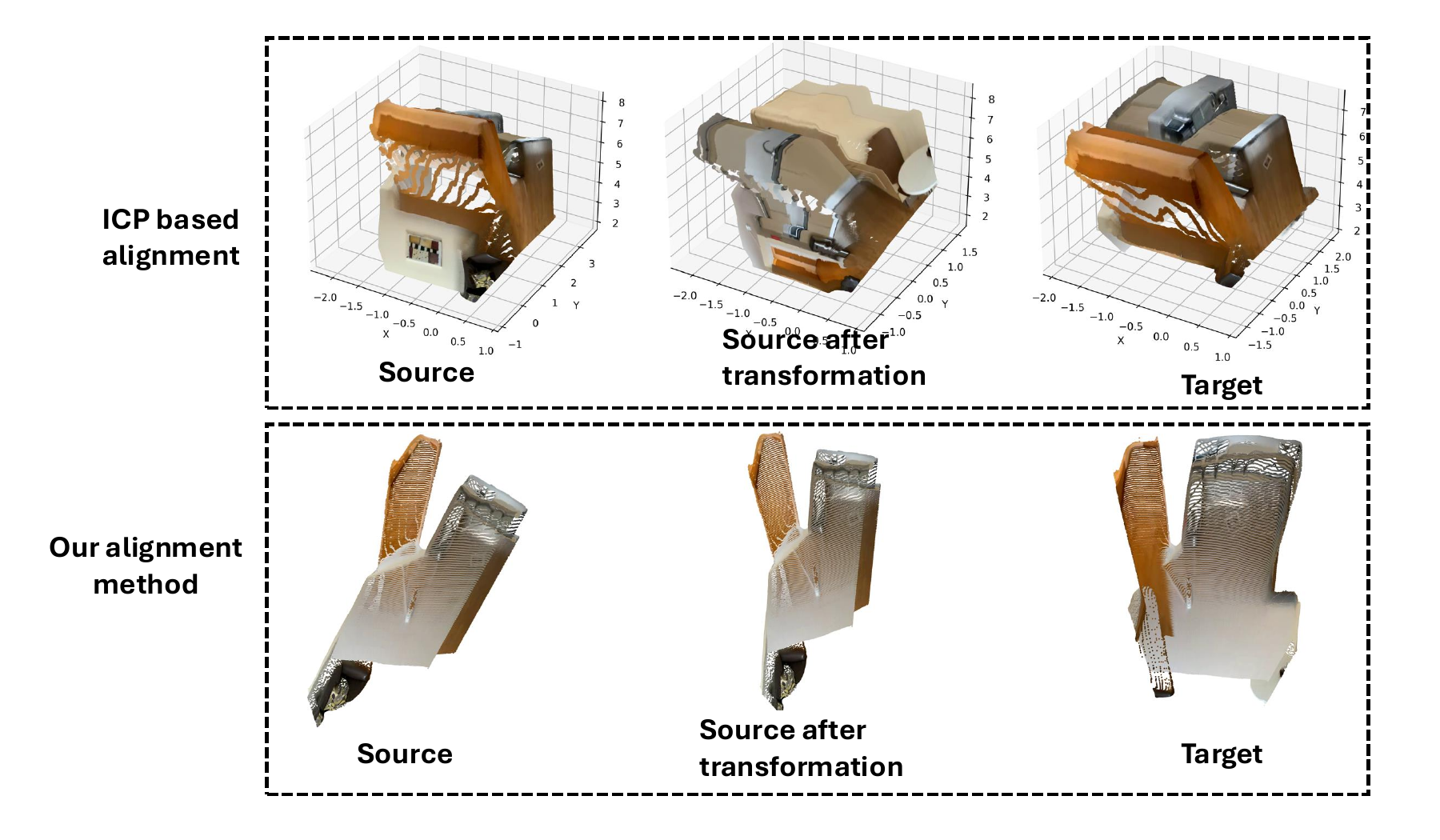}
	\vspace{-0.3in}
	\caption{Comparison between ICP based alignment and our alignment emthod}
	\label{fig:different_alignment}
	\vspace{-0.15in}
\end{wrapfigure}

\section{Reasoning with Iterative Visual Correlations}

\subsection{Cross-Frame alignment}
\label{subsec:cross_frame_alignment_details}

As described in Section \ref{subsec:iterative_construction}, to efficiently compute the transformation matrix, we adopt deep feature-based image matching models to establish pixel correspondences in RGB images and subsequently align the 3D coordinates of the matched points. To further reduce alignment error, rather than using all pixel correspondences identified by the image matching model, we retain only those within the pre-computed bounding boxes of task-related objects, as shown in Figure \ref{fig:matched_pixels}. In this way, as shown Figure \ref{fig:different_alignment}, we can effectively avoid involving irrelvant background regions such as walls and floors, hence ensuring accurate alignment. 

However, image matching models can only process a single pair of frames per inference. If each unaligned frame is matched against all previously aligned frames, the computational cost would grow quadratically with the number of frames. To further improve the compute efficiency, we instead match each unaligned frame to the most similar aligned frames. 
Assuming that the same target object maintains consistent structural patterns when captured from different viewpoints, we employ lightweight similarity measures such as PSNR or SSIM \cite{wang2004image}, and results in Table \ref{tab:sim_metric} show that accuracy of alignment using these simple metrics well approximate that of using expensive neural-based metrics such as FID \cite{heusel2017gans,zhang2018unreasonable}.

\begin{table}[h]
    \small
	\centering
	\begin{tabular}{lccc}
		\hline
		\textbf{} & \textbf{PSNR} & \textbf{SSIM} & \textbf{Hist}\\
		\hline
		\textbf{Selection Acc (\%)}& 95.5 & 97.3 &94.2\\
		
		\hline
	\end{tabular}
\vspace{0.05in}
	\caption{Performance of alignment, compared to that of using FID \cite{heusel2017gans} as the similarity metric}
		\vspace{-0.05in}
	\label{tab:sim_metric}
\end{table}



\subsection{Multi-Frame Alignment}
\label{subsec:multi_frame_alignment}
To integrate these pairwise alignments into a globally consistent map, a naive approach is sequential integration, which iteratively estimates the transformation $\mathbf{T}_{t \to t+1}$ between consecutive time steps. Under this scheme, the global coordinates of a point $\mathbf{p}_t$ are derived by chaining transformations:
\begin{equation}
\mathbf{p}_{t+k}^{\text{global}} = \mathbf{T}_{t+k-1 \to t+k} \cdots \mathbf{T}_{t \to t+1} \mathbf{p}_t.
\end{equation}

\begin{wrapfigure}{r}{3.3in}
	\centering
	\vspace{-0.1in}
	\includegraphics[width=\linewidth]{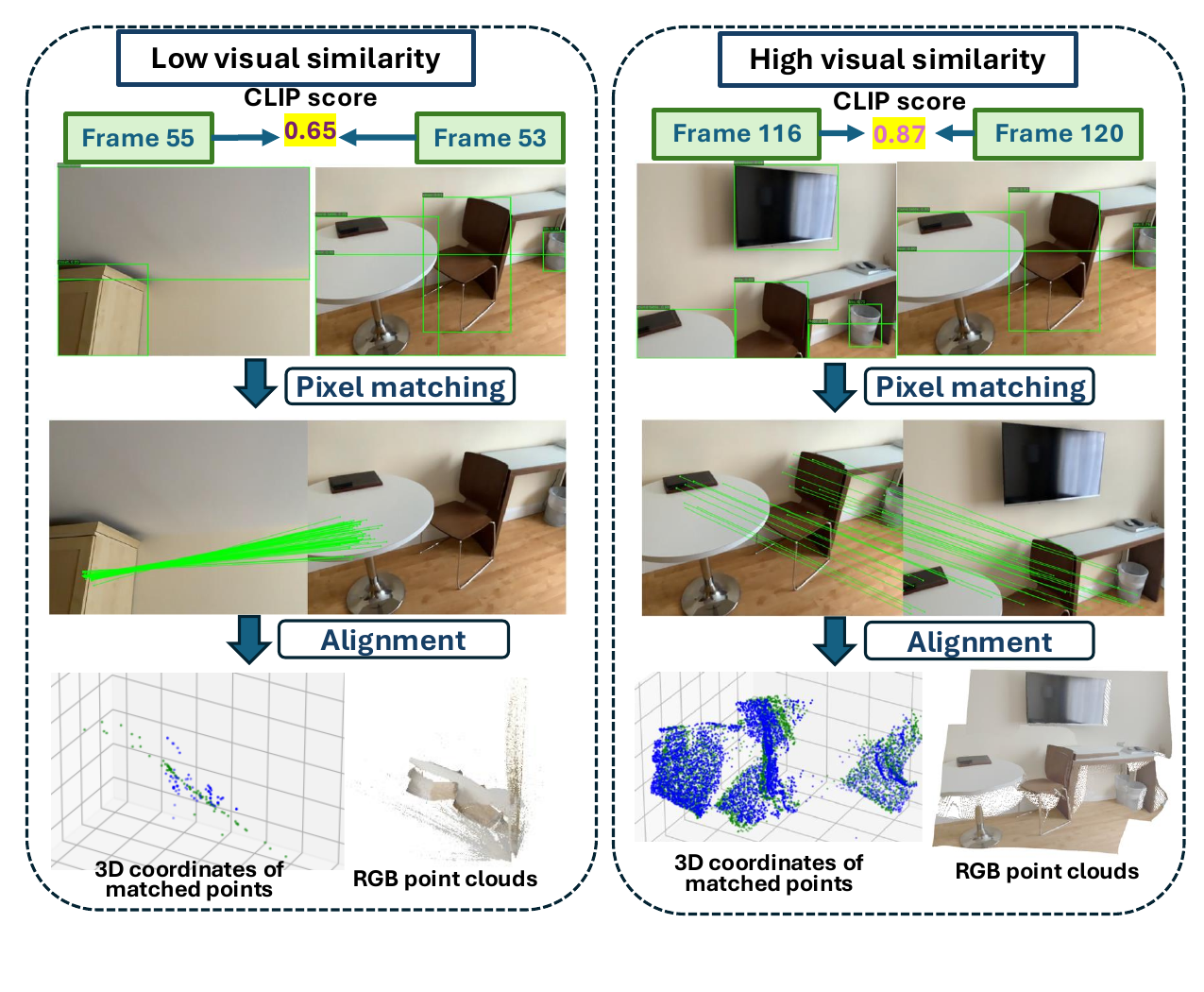}
	\vspace{-0.45in}
	\caption{Comparison of pairwise alignment results with different visual similarities}
	\label{fig:pairwise_matches}
	\vspace{-0.1in}
\end{wrapfigure}

However, this sequential method suffers from two critical limitations. First, as shown in the left side of Figure \ref{fig:pairwise_matches}, temporal adjacency of frames in a sparsely sampled video does not guarantee sufficient visual similarity for accurate registration. If the camera moves significantly between two selected frames, the transformation estimation may fail. Second, this approach is highly susceptible to the cumulative error (drift). A single inaccurate estimation can propagate the estimation error through the transformation chain, causing the global representation of subsequent frames to diverge and an incoherent semantic map being rendered.

To overcome these limitations, we propose a topology-aware alignment strategy. Instead of integrating pairwise alignments of frames sequentially, we organize the video frames into a similarity tree structure, maximizing the visual overlap between connected nodes, each of which represents a frame. This formulation ensures that global alignment relies exclusively on the estimates between visually similar frames with sufficient feature overlap. Furthermore, this topology provides inherent error isolation, because frames that lack similarity with the all other frames naturally correspond to leaf nodes in the tree. As a result, any estimation errors associated with these outliers remain localized and do not propagate to corrupt the alignment of other frames.



Specifially, let $\mathcal{V} = \{1, \dots, N\}$ be the set of indices for the selected key frames. We define a fully connected similarity graph $\mathcal{G} = (\mathcal{V}, \mathcal{E})$. To minimize computational overhead, the weight $w_{ij}$ of an edge $(i, j) \in \mathcal{E}$ is defined as the cosine similarity between the CLIP embeddings of frame $i$ and frame $j$. This metric provides a computationally efficient proxy for semantic overlap prior to detailed geometric matching. We then construct a \textit{Tree} $\mathcal{T}$ from $\mathcal{G}$ to determine the optimal alignment path. This process consists of three steps, as described below:

\vspace{0.05in}
\noindent\textbf{1. Root Selection.} We identify a global anchor frame (root), denoted as $r$, which serves as the origin of the global coordinate system. The root is selected as the frame with the highest aggregate similarity to all other frames, ensuring maximal centrality:
\begin{equation}
    r = \operatorname*{argmax}_{i \in \mathcal{V}} \sum_{j \in \mathcal{V}, j \neq i} w_{ij}.
\end{equation}

\vspace{0.05in}
\noindent\textbf{2. Tree Construction.} To ensure that transformations are estimated exclusively between visually similar pairs, we construct $\mathcal{T}$ as the Maximum Spanning Tree (MST) of $\mathcal{G}$. The tree topology is optimized to maximize the sum of edge weights:
\begin{equation}
    \mathcal{T} = \operatorname*{argmax}_{\mathcal{T}' \subseteq \mathcal{G}} \sum_{(i, j) \in \mathcal{E}(\mathcal{T}')} w_{ij},
\end{equation}

\begin{wrapfigure}{r}{2.6in}
	\centering
	\vspace{-0.1in}
	\includegraphics[width=\linewidth]{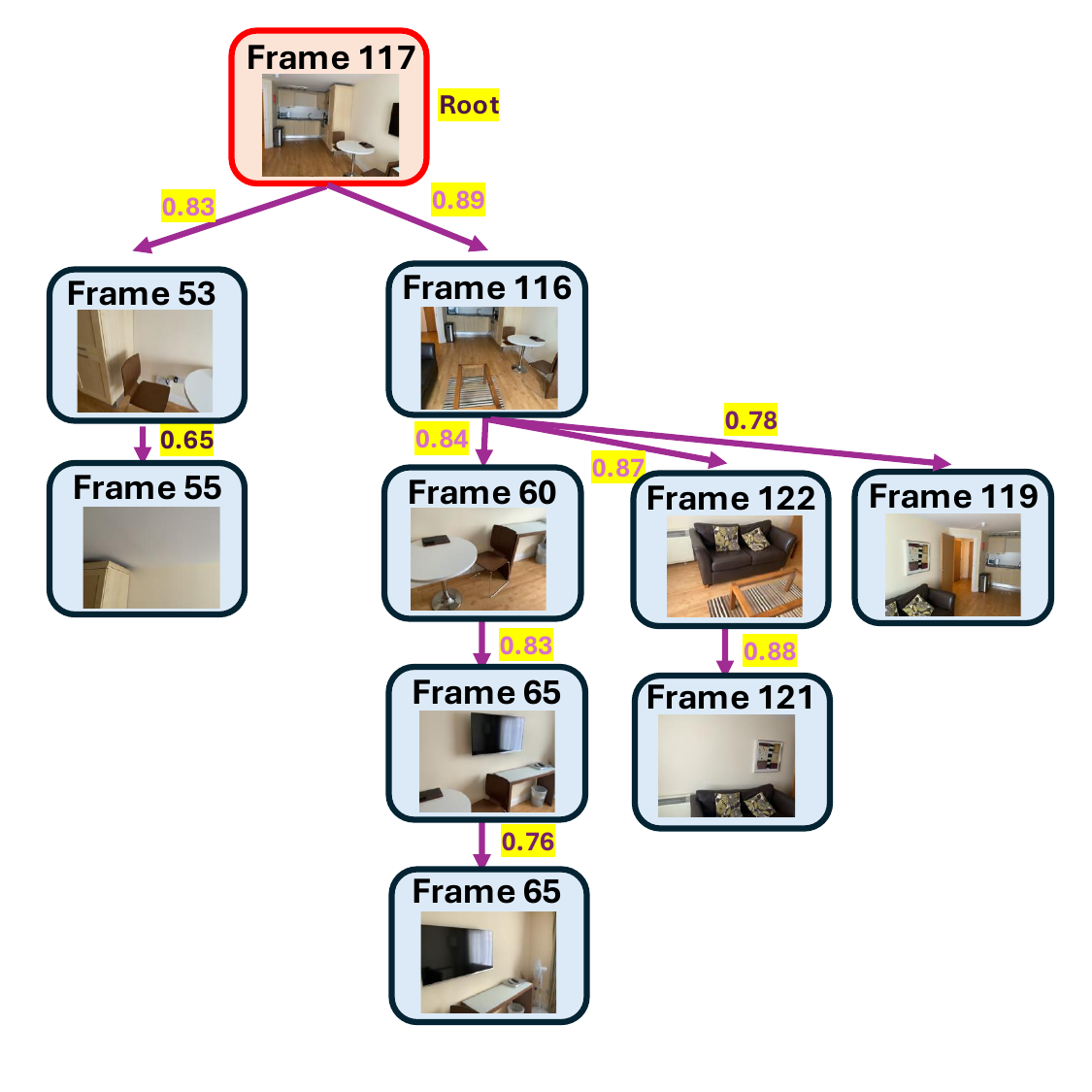}
	\vspace{-0.4in}
	\caption{An example of the constructed frame similarity tree}
	\label{fig:spm_tree}
	\vspace{-0.05in}
\end{wrapfigure}

where $\mathcal{E}(\mathcal{T}')$ denotes the edge set of a candidate tree. We implement this optimization using Kruskal's algorithm ($O(N^2 \log N)$), which iteratively adds the edges with the highest similarity scores that do not form cycles. This structure ensures that every frame connects to the graph via its most reliable visual connections, while frames with low overall similarity are naturally pushed to the leaves of the tree. Figure \ref{fig:spm_tree} illustrates an example of the constructed similarity tree, where Frame 55 is positioned as a leaf node due to its low similarity with other frames. If we utilize the naive sequential alignment scheme, an inaccurate transformation estimation between Frames 53 and 55 would propagate errors to all subsequent frames; our tree topology prevents this cascade.

\vspace{0.05in}
\noindent\textbf{3. Global Transformation Computation.} For any frame $i$, there exists a unique path in $\mathcal{T}$ connecting it to the root $r$. Let this path be represented as a sequence of nodes $P_{i \to r} = (v_0, v_1, \dots, v_m)$, where $v_0 = i$ and $v_m = r$. The global transformation matrix $\mathbf{M}_i \in SE(3)$, which maps points from the local frame $i$ to the global coordinate system (frame $r$), is computed by compounding the pairwise transformations along this path:
\begin{equation}
    \mathbf{M}_i = \mathbf{T}_{v_{m-1} \to v_m} \cdot \mathbf{T}_{v_{m-2} \to v_{m-1}} \cdots \mathbf{T}_{v_0 \to v_1},
\end{equation}
where $\mathbf{T}_{u \to v}$ is the relative transformation matrix derived from image matching between adjacent nodes $u$ and $v$ in the tree. By strictly following the path of maximum similarity, this formulation isolates errors from outlier frames at the branches' ends and prevents drift propagation.

\begin{wrapfigure}{r}{2.7in}
	\centering
		\vspace{-0.1in}
	\includegraphics[width=\linewidth]{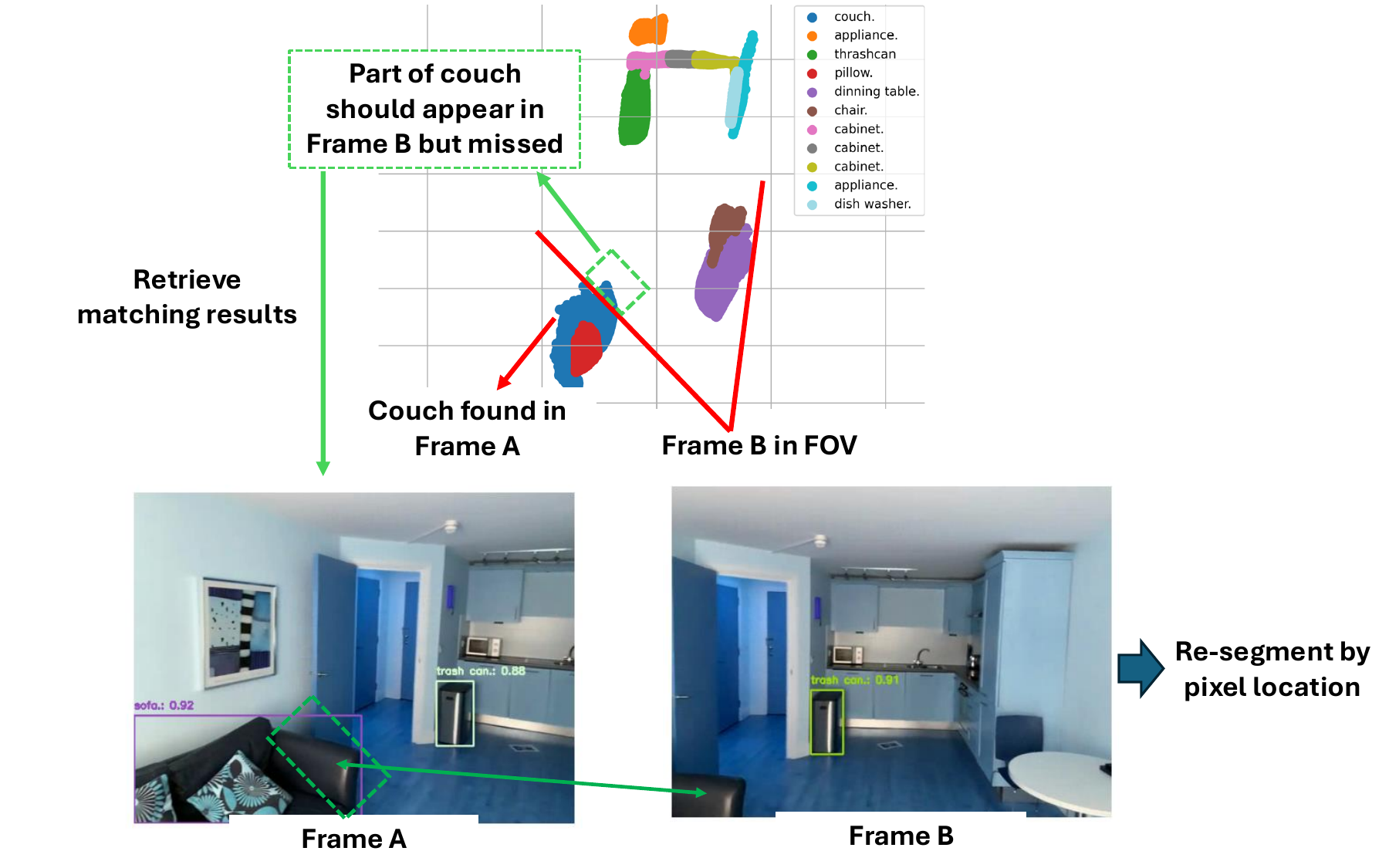}
	\vspace{-0.25in}
	\caption{Refinement stage for missing objects}
	\label{fig:final_refinement}
	\vspace{-0.1in}
\end{wrapfigure}

\subsection{Tackling Occlusion and Partial Visibility of Objects}
\label{subsec:final_refinement}

To address occlusion and partial visibility of objects in the scene, our approach is to introduce an extra refinement stage following the completion of cross-frame alignment. Once the transformation matrices are calculated, 
we use the calibrated camera intrinsics and extrinsics to identify objects that are geometrically expected to fall within the current frame's FoV but are missing from the segmentation results. In such cases, cross-frame pixel correspondences established during earlier alignment steps provide candidate locations for the undetected object. These pixel locations, then, are used to guide the segmentation model, to recover the partially occluded or fragmented objects. 

For example, as illustrated in Figure \ref{fig:final_refinement}, if an object is successfully segmented in Frame A, then based on the constructed scene and estimated camera pose, its corresponding pixels should also appear in Frame B. If the object is not detected in Frame B, we search for pixel correspondences between the segmented region in Frame A and the unsegmented regions in Frame B. The matched pixels are subsequently used as priors to guide the segmentation model to recover the missing object in Frame B.

\section{Optimizing the Selection of Key Frames}	
To make the semantic map construction and the spatial reasoning more computationally efficient, as described in Section 3.3, we introduce a key frame selection framework, which aims to identify a minimal yet sufficient subset of frames that maximizes the coverage of spatial information for the specific reasoning task. We first propose a Gaussian kernel-based update mechanism that leverages the temporal locality of object appearance to progressively refine the sampling distribution around high-confidence detections (\cref{subsec:gaussian_kernel}). Afterwards, we analyze critical design choices that optimize the trade-off between the frame selection accuracy and on-device computational efficiency (\cref{subsec:selection_design_choice}).

\subsection{Applying the Gaussian Kernel}
\label{subsec:gaussian_kernel}

When updating the sampling distribution of key frame selection, we enforce the temporal locality by applying a Gaussian kernel to each scored frame, so that other frames that are temporally close to this scored frames will have higher chances to be scored. More specifically, letting the input video frames be indexed as $1, 2, ... N$ in time, the Gaussian kernel applied to a scored frame $i$ can be written as
\begin{equation}
\textsc{GaussianKernel}(x;\, I_i,\sigma_i^2)
= \exp\!\left(-\frac{(x - I_i)^2}{2\sigma_i^2}\right),
\end{equation}
where $I_i$ is the likelihood score of frame $i$, and the kernel width $\sigma_i$ (i.e., the standard deviation) is adaptively determined by the average normalized similarity score ($s_i$), measured by histogram-based metrics \cite{swain1991color, zhao2019key}, between frame $i$ and other frames within a time window centered at $i$. In practice, the proper size of time window depends on the specific characteristics of temporal locality between frames, which is scene dependent. To further verify the temporal locality and quantize such time window, we conducted a small-scale benchmark study using the VSI-Bench benchmark that depicts indoor scenes \cite{yang2025thinking}. Based on results in Appendix \ref{appendix:temporal_locality}, we set the size of time window ($r$) as 3, to ensure that the Gaussian kernel can capture short-range temporal correlations while avoiding other frames with weak semantic relevance.

To ensure alignment between the kernel spread and the window, we define the minimum and maximum standard deviations as
\begin{equation}
\sigma_{\min} = \frac{1}{k}, \qquad \sigma_{\max} = \frac{r}{k},
\end{equation}
where \(k = \sqrt{2}\,\operatorname{erf}^{-1}(I_i; \alpha)\) and \(\alpha \in (0, 1)\) specifies the desired fraction of probability mass within \([I_i-k\sigma_i, I_i+k\sigma_i]\). For example, setting \(\alpha = 0.6827\) recovers \(k \approx 1\) (the \(1 \sigma_i\) band), while \(\alpha = 0.9545\) yields \(k \approx 2\) (the \(2 \sigma_i\) band). Given these bounds, we interpolate the variance as
\begin{equation}
\sigma_i(s_i) \;=\; \sigma_{\min} \;+\; \bigl(\sigma_{\max} - \sigma_{\min}\bigr)\,s_i, \quad s_i \in [0, 1]
\end{equation}

By setting \(k=2\), we obtain \(\sigma_i(s_i) \in \bigl[\tfrac{1}{2},\, \tfrac{r}{2}\bigr]\), which ensures that when frame similarity is low in the  window, the kernel collapses towards a narrow peak around $i$ within the adjacent frame index, restricting influence to adjacent frames; conversely, when the frame similarity is high, the kernel allocates most of its mass across the entire window, facilitating evidence propagation among semantically similar frames.



\subsection{Design Choices in Key Frame Selection}
\label{subsec:selection_design_choice}

Principally, MosaicThinker only selects a small set of key frames for constructing the semantic map. In practice, we pre-define the number of key frames before selection, based on the available on-device compute power and the required latency of reasoning. For example, in scenarios of daily assistance with robotic helpers, the end-to-end latency should be within 1 second \cite{liu2019edge, satyanarayanan2017emergence}. If it takes 40ms to decide if a frame contains task-related objects, we can have 25 key frames for reasoning, and more key frames can be involved on stronger embodied AI platforms such as autonomous vehicles.

\vspace{0.05in}
\noindent\textbf{Threshold of likelihood score.} As we described in Section \ref{subsec:keyframe_selection}, a frame is selected as the key frame if its likelihood score of containing task-related objects is higher than a threshold, which controls the tradeoff between recall and precision in key frame selection: a low threshold tends to include more key frames, possibly improving the coverage of task-related objects and also risking redundancy. Conversely, a high threshold makes it more difficult to select key frames, and also more likely to miss frames that contain task-related objects with occlusion or low confidence.

\vspace{-0.05in}
\begin{table}[ht]
	\centering
	\small
	\setlength{\tabcolsep}{4pt}
	\begin{tabular}{l c ccc ccc}
		\toprule
		{Method} & Threshold  & Precision & Recall & $F_1$ \\
		\midrule
		Uniform sampling & -  & 0.29 & 0.33  & 0.31 & \\
		Retrieval-based & -  & 0.61 & \textbf{0.79} & 0.69 & \\
		VideoAgent & -  & 0.58 & 0.66  & 0.63 & \\
		Iterative search (ours) & 1  & 0.55 & 0.78  & 0.67 & \\
		Iterative search (ours)  & \textbf{1.5}  & 0.65 & 0.76 & \textbf{0.70} & \\ 
		Iterative search (ours)  & 2  & \textbf{0.66} & 0.67  & 0.66 & \\ 
		\bottomrule
	\end{tabular}
\vspace{0.05in}
	\caption{The impact of different values of the likelihood score threshold in key frame selection}
	\vspace{-0.0in}
	\label{tab:threshold}
\end{table}

To further quantify the impact of this threshold on key frame selection, over the LV-HAYSTACK dataset that contains 1,102 video clips with a total length of 450 hours \cite{ye2025re}. Results in Table \ref{tab:threshold} show that this threshold critically shapes the aforementioned tradeoff as we expect. With all values of threshold, our approach of iterative temporal search constantly approaches the performance of the naive Retrieval-based method which scores and ranks all frames for frame selection, and outperforms naive sampling methods, including (1) Uniform sampling, which selects frames at fixed intervals; and 2) LLM-based search methods such as VideoAgent \cite{wang2024videoagent}, which leverage LLM for key frame selection in videos. In practice, this advantage allows users to flexibly decide the specific value of threshold based on the application requirement, and a medium value of 1.5 generally balances between aspects of recall and precision.

\vspace{-0.05in}
\begin{table}[ht]
	\centering
	\small
	\setlength{\tabcolsep}{6pt}
	\begin{tabular}{c c c c c c}
		\toprule
		\# of iterations & Precision & Recall & $F_1$ & Sampled Frame (\%) \\
		\midrule
		100  & 0.57 & 0.66 & 0.63& 10.2 \\
		33  & 0.62 & 0.71 & 0.66 & 25.6 \\
		\textbf{20}  & \textbf{0.65} & \textbf{0.76} & \textbf{0.70}& 26.1 \\ 
		14  & 0.65 & 0.72 & 0.68 & 30.7 \\ 
		11  & 0.63 & 0.70 & 0.67 & 37.2 \\ 
		\bottomrule
	\end{tabular}
\vspace{0.05in}
	\caption{Performance of key frame selection with different numbers of iterations in temporal search}
	\vspace{-0.2in}
	\label{tab:batch_size}
\end{table}

\vspace{0.1in}
\noindent\textbf{Number of iterations.} The number of iterations in temporal search determines the granularity of key frame selection. More iterations updates the sampling distribution more frequently and hence improves adaptivity, but more iterations also incur higher compute costs. Conversely, fewer iterations is more computationally efficient, but increases the chance of involving redundancy in key frames. As shown in Table~\ref{tab:batch_size} where we tested the performance of key frame selection on the LV-HAYSTACK dataset, a moderate number of iterations (e.g., 20 iterations) achieves the performance in precision, recall and F1 score.


\section{Experiment Settings}
\label{sec:experiment_setting}
We deployed MosaicThinker on various types of mobile and embedded AI platforms, including NVidia Jetson Orion and Oneplus Ace 3 Pro smartphone. We then select VLMs with various parameter sizes to match the compute power constraints of each device: Qwen3-VL-32B-4bit \cite{bai2025qwen2} and InternVL3-8B \cite{chen2024expanding} on NVidia Jetson Orion and Qwen-2.5-VL-3B \cite{bai2025qwen2} on the smartphone.

\noindent\textbf{System setup.} On NVidia Jetson Orion, VLM inference was operated using PyTorch with TensorRT acceleration, enabling optimized execution on the device GPU. This configuration leverages mixed-precision computation to reduce latency and improve throughput while maintaining accuracy. To deploy VLMs and run VLM inference on the smartphone, we used the ONNX Runtime \cite{onnx} for Android, leveraging its NNAPI backend to enable hardware acceleration using the device's heterogeneous computing units. This setup automatically switches between GPU and CPU execution based on workload characteristics, ensuring optimal performance while maintaining energy efficiency. 

\begin{wrapfigure}{r}{3in}
	\centering
		\vspace{-0.0in}
	\includegraphics[width=\linewidth]{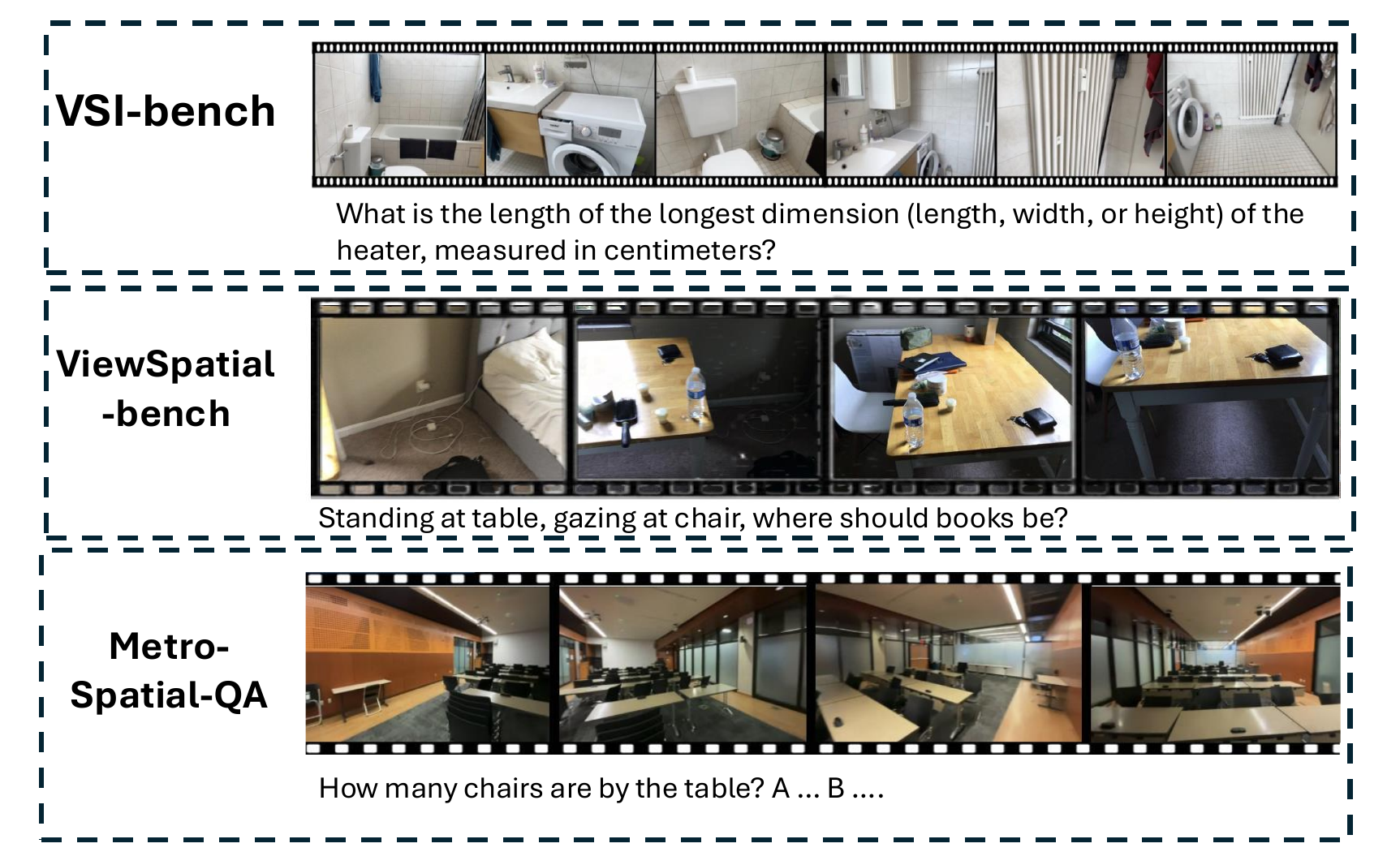}
	\vspace{-0.3in}
	\caption{Example of data samples from benchmark datasets}
	\label{fig:bench_eg}
\vspace{-0.0in}
\end{wrapfigure}

\noindent\textbf{Benchmarks.} 
Our evaluations use three categories of spatial reasoning tasks, including 1) \emph{Metric Measurement} (encompassing absolute/relative distance, object size, and object location) to verify the model's understanding of scale and depth; 2) \emph{Perspective-Dependent Reasoning} (covering relative direction and object counting) to test the ability to perform ``mental rotation''; and 3) \emph{Spatiotemporal Tracking} (identifying the appearance order of objects in video) to assess whether the model can maintain a persistent memory of object locations. These tasks have been widely considered as representing the most fundamental spatial reasoning capabilities that are transferrable to real-world application scenarios \cite{brown2025sims}. 

In our experiments, 
we selected multiple benchmark datasets that cover these three categories of tasks across various room types. Details of these benchmark datasets are described as below, and images of data samples in these benchmarks are in Figure \ref{fig:bench_eg}.
\vspace{-0.05in}
\begin{itemize}
	\item \textbf{VSI-Bench \cite{yang2025thinking}:} This benchmark comprises over 5,000 question–answer (QA) pairs from 288 real-world videos of indoor scenes. It includes 8 question types, of which we select 5 that are the most relevant to spatial reasoning for on-device embodied AI: absolute distance (A. D.), object size (O. S.), relative distance (R. Dt), object count (O. C.), relative direction (R. Dr.) and appearance order (Ap. Or.). 
    \vspace{0.05in}
	\item \textbf{STI-Bench \cite{li2025sti}:} This benchmark contains 300 videos and over 2,000 QA pairs designed for spatio-temporal video understanding. It includes 8 question types, of which we select 2 for spatial reasoning: dimensional measurements (Dim. Mea.), and 3D video grounding (Grd.). 
	\vspace{0.05in}
    \item \textbf{Metro-Spatial-QA}: The above two benchmarks cover only limited room types (residential homes and offices). To address this gap, we built a new dataset, namely \emph{Metro-Spatial-QA}, which comprises 40 egocentric video clips that are recorded in 5 diverse indoor settings, including supermarkets, libraries, museums, restaurants and classrooms. These video clips are then paired with two types of manually annotated spatial reasoning tasks: measurements and perspective taking, yielding 160 spatial reasoning QA samples (4 questions for each video). Since these scenes are specifically designed to capture the complexity of real-world activities, they feature challenging variables such as dynamic object distributions, varying lighting conditions, and frequent occlusions, which provide new aspects of evaluating visual spatial reasoning and cross-frame understanding. More details are in Appendix \ref{appendix:dataset}.
\end{itemize}

In these benchmarks, most spatial reasoning tasks are formulated as multi-choice questions, and we are hence able to evaluate the accuracy of spatial reasoning as the percentage of correct answer choices. For some other types of reasoning tasks such as deciding the absolute distance and object size, the answers are numerical and we instead evaluate the performance of spatial reasoning using the Mean Relative Accuracy.

\noindent\textbf{Baselines.} Since MosaicThinker is designed as an inference-time technique that does not require any model training efforts, we exclude other training-based spatial reasoning methods \cite{chen2024spatialvlm,chen2024ll3da,yang2025cambrian,hong20233d} from evaluation, but instead compared MosaicThinker against against 5 representative training-free baselines to evaluate its effectiveness in enhancing spatial reasoning capability of off-the-shelf VLMs. These baselines are described as below:
\vspace{-0.15in}
\begin{itemize}
	\item \textbf{Direct input}: It is a naive baseline that directly downsamples the video and feeds the sampled video frames into the VLM for spatial reasoning. 
    \vspace{0.05in}
	\item \textbf{Video-COT prompting \cite{hu2025cos}}: It is a prompting method for video understanding, which treats the selection of video shots as an optimization problem. It uses a binary video summary to identify task-relevant shots and a co-reasoning module to pair these relevant shots with irrelevant ones, effectively guiding the model to focus on the most important content.
    \vspace{0.05in}
	\item \textbf{Scene graph reconstruction \cite{chu2025understanding}}: It models long videos with an entity–relation graph, where an LLM tracks entities and their evolving interactions over time.
    \vspace{0.05in}
	\item \textbf{APC (Mental imagery simulation)
\cite{lee2025perspective}}: It is a framework that enables VLMs to reason from arbitrary viewpoints by converting a 2D image into a coarse 3D abstraction, mathematically transforming that abstraction into a target coordinate frame, and prompting the model with the resulting perspective-aware data.
    \item \textbf{Ground truth semantic map}: With VSI-Bench and STI-Bench benchmarks, we also evaluated the VLM performance using ground truth semantic maps, which are derived from the 3D point clouds and segmentation masks included in the indoor scene datasets they used (e.g., ScanNet \cite{dai2017scannet}). This serves as an explicit validation of the semantic maps generated by MosaicThinker; a smaller performance gap indicates higher accuracy and fewer errors in our constructed maps. 
\end{itemize}

Note that, when we use the ground truth semantic map in spatial reasoning, we achieve the best spatial reasoning performance that can be possibly achieved among using all types of spatial information input. In practice, even in this case, VLMs could still make significant errors in spatial reasoning, especially when VLMs with small parameter sizes are used for on-device deployment. However, these errors arise from limitations of VLMs' model architecture and representation power, which call for model redesign that is out of the scope of this paper. Our focus, instead, is to validate that MosaicThinker can effectively approximate to such best possible performance by providing sufficiently accurate semantic map via iterative integration and construction.

\begin{table*}[ht]
\centering

\resizebox{\textwidth}{!}{
\vspace{-0.3in}
\begin{tabular}{lccccccccccccc}
\toprule
\multirow{2}{*}{\textbf{Method}} & \multicolumn{7}{c}{\textbf{VSI-Bench}} & \multicolumn{3}{c}{\textbf{STI-Bench}} & \multicolumn{3}{c}{\textbf{Metro-Spatial-QA}} \\ 
\cmidrule(lr){2-8} \cmidrule(lr){9-11} \cmidrule(lr){12-14}
 & A.D. & O.S. & R.Dt. & R.Dr & O.C. & Ap.Or. & \textbf{Avg} & D.Meas. & Grd. & \textbf{Avg.} & Meas. & Pers. & \textbf{Avg.}\\ 
\midrule
Direct Input                       & 27.3 & 35.1 & 35.9 & 33.2 & 23.5 & 30.1 & 30.2 & 25.8 & 27.4 & 26.2 & 37.3 & 37.5 & 37.4 \\
Video-CoT                 & 25.9 & 31.9 & 36.8 & 31.5 & 22.8 & 28.6 & 28.7 & 26.5 & 27.9 & 27.8 & 37.8 & 38.3 & 38.1 \\
Scene Graph Reconstruction               & 27.4 & 35.0 & 37.1 & 33.7 & 23.3 & 28.7 & 29.8 & 27.6 & 28.4 & 28.0 & 38.4 & 38.1 & 38.2 \\
APC                       & 26.8 & 34.4 & 37.0 & 34.4 & 23.0 & 29.5 & 31.1 & 29.1 & 30.2 & 29.6 & 40.7 & 38.9 & 39.3 \\
MosaicThinker           & 29.4 & 36.6 & 38.9 & 33.5 & 28.3 & 32.2 & 33.4 & 30.8 & 31.8 & 31.6 & 45.5 & 40.1 & 42.8 \\
\midrule
Ground Truth Semantic Map & 30.3 & 37.7 & 39.5 & 34.0 & 31.7 & 32.4 & 34.6 & 31.3 & 32.9 & 32.1 & - & - & - \\
\bottomrule
\end{tabular}
}
\caption{Performance comparison on the Oneplus 12R Smartphone with Qwen-2.5-VL-3B}
\vspace{-0.35in}
\label{tab:main_3B}
\end{table*}

\section{Performance Evaluation}

\subsection{Spatial Reasoning Accuracy}

We conduct evaluations on three benchmarks with various
quantitative spatial reasoning tasks, covering
small, medium, and relatively large vision-language backbones that can be deployed on embodied AI devices. The
results are in Table \ref{tab:main_3B}, Table \ref{tab:main_8B}, and Table \ref{tab:main_32B}, respectively. 
Evaluations across models of different sizes confirm that MosaicThinker consistently enhances spatial reasoning accuracy, with different VLMs and benchmark settings. With larger VLMs on the MetroSpatial-QA benchmark, such accuracy enhancement can be up to 11\%. In comparison, baselines like Video-CoT and Scene Graph Reconstruction show minimal improvement; lacking explicit 3D information, their extended context windows often overwhelm the model's reasoning capabilities, occasionally degrading performance. While APC incorporates 3D data, it presents it frame-by-frame, and current models struggle to integrate such fragmented information, resulting in only marginal gains. In contrast, MosaicThinker excels by stitching multiview data into a coherent global semantic map, driving significant improvements in tasks like object counting and relative distance estimation where single-frame visibility is limited. 

Although a small performance gap still remains compared to the ``perfect'' ground-truth map, due to depth estimation noise, the oracle's superior results validate our core hypothesis: providing VLMs with an explicit, unified 3D representation is essential for achieving better and robust spatial intelligence.

\begin{table*}[t]
\centering

\resizebox{\textwidth}{!}{
\vspace{-0.1in}
\begin{tabular}{lccccccccccccc}
\toprule
\multirow{2}{*}{\textbf{Method}} & \multicolumn{7}{c}{\textbf{VSI-Bench}} & \multicolumn{3}{c}{\textbf{STI-Bench}} & \multicolumn{3}{c}{\textbf{Metro-Spatial-QA}} \\ 
\cmidrule(lr){2-8} \cmidrule(lr){9-11} \cmidrule(lr){12-14}
 & A.D. & O.S. & R.Dt. & R.Dr & O.C. & Ap.Or.  & \textbf{Avg} & D.Meas. & Grd. & \textbf{Avg.} & Meas. & Pers. & \textbf{Avg.}\\ 
\midrule
Direct Input                       & 32.2 & 44.5 & 42.8 & 37.7 & 67.8 & 47.2 & 42.0 & 28.5 & 34.9 & 31.8 & 40.1 & 41.2 & 40.7  \\
Video-CoT                 & 31.9 & 43.7 & 42.7 & 38.6 & 68.5 & 46.1 & 44.1 & 29.1 & 35.0 & 32.5 & 41.5 & 41.6 & 41.5  \\
Scene Graph Reconstruction               & 32.4 & 44.5 & 42.4 & 38.5 & 70.7 & 46.5 & 43.7 & 29.2 & 35.1 & 32.2 & 42.2 & 42.3 & 42.2 \\
APC                       & 32.0 & 43.8 & 42.0 & 39.4 & 67.1 & 47.1 & 46.8 & 31.1 & 35.7 & 34.4 & 44.8 & 46.1 & 45.4 \\
MosaicThinker           & 36.6 & 51.2 & 48.2 & 39.7 & 76.1 & 49.3 & 50.8 & 35.3 & 38.9 & 38.1 & 51.9 & 50.7 & 51.2  \\
\midrule
Ground Truth Semantic Map & 37.3 & 51.6 & 49.9 & 40.2 & 85.6 & 49.2 & 52.9 & 36.8 & 39.0 & 38.3 & - & - & -  \\
\bottomrule
\end{tabular}
}
\caption{Performance comparison on NVidia Jetson Orion with InternVL3-8B}
\label{tab:main_8B}
\end{table*}

\begin{table*}[t]
\centering

\resizebox{\textwidth}{!}{
\vspace{-0.1in}
\begin{tabular}{lccccccccccccc}
\toprule
\multirow{2}{*}{\textbf{Method}} & \multicolumn{7}{c}{\textbf{VSI-Bench}} & \multicolumn{3}{c}{\textbf{STI-Bench}} & \multicolumn{3}{c}{\textbf{Metro-Spatial-QA}} \\ 
\cmidrule(lr){2-8} \cmidrule(lr){9-11} \cmidrule(lr){12-14}
 & A.D. & O.S. & R.Dt. & R.Dr & O.C. & Ap.Or. & \textbf{Avg} & D.Meas. & Grd. & \textbf{Avg.} & Meas. & Pers.  & \textbf{Avg.}\\ 
\midrule
Direct Input                       & 47.7 & 76.3 & 58.7 & 47.2 & 67.4 & 44.1 & 56.4 & 29.5 & 32.2 & 30.8 & 45.6 & 47.3 & 45.9  \\
Video-CoT                 & 46.1 & 77.4 & 59.7 & 47.9 & 68.4 & 44.9 & 57.7 & 30.0 & 31.8 & 30.9 & 44.2 & 48.0 & 46.1  \\
Scene Graph Reconstruction               & 44.5 & 76.5 & 59.1 & 48.0 & 70.2 & 42.7 & 58.6 & 30.7 & 31.9 & 31.3 & 44.7 & 49.2 & 47.3  \\
APC                       & 48.7 & 75.9 & 62.2 & 49.4 & 66.3 & 43.3 & 62.5 & 32.0 & 30.8 & 31.4 & 45.4 & 50.6 & 48.2  \\
MosaicThinker           & 53.2 & 79.3 & 64.4 & 51.8 & 77.2 & 45.0 & 67.1 & 36.4 & 33.9 & 34.8 & 49.2 & 54.0 & 52.6  \\
\midrule
Ground Truth Semantic Map & 55.2 & 80.5 & 65.1 & 53.6 & 87.4 & 45.2 & 68.3 & 37.5 & 34.4 & 35.9 & - & - & -  \\
\bottomrule
\end{tabular}
}
\caption{Performance comparison on Jetson with Qwen3VL-32B-4bit}
\label{tab:main_32B}
\end{table*}

\subsection{On-device Compute Efficiency}
\begin{figure}[ht]
    \centering
    \hspace{-0.25in}
    \subfigure[Qwen-2.5-VL-3B on Smartphone]{
        \includegraphics[width=0.35\linewidth]{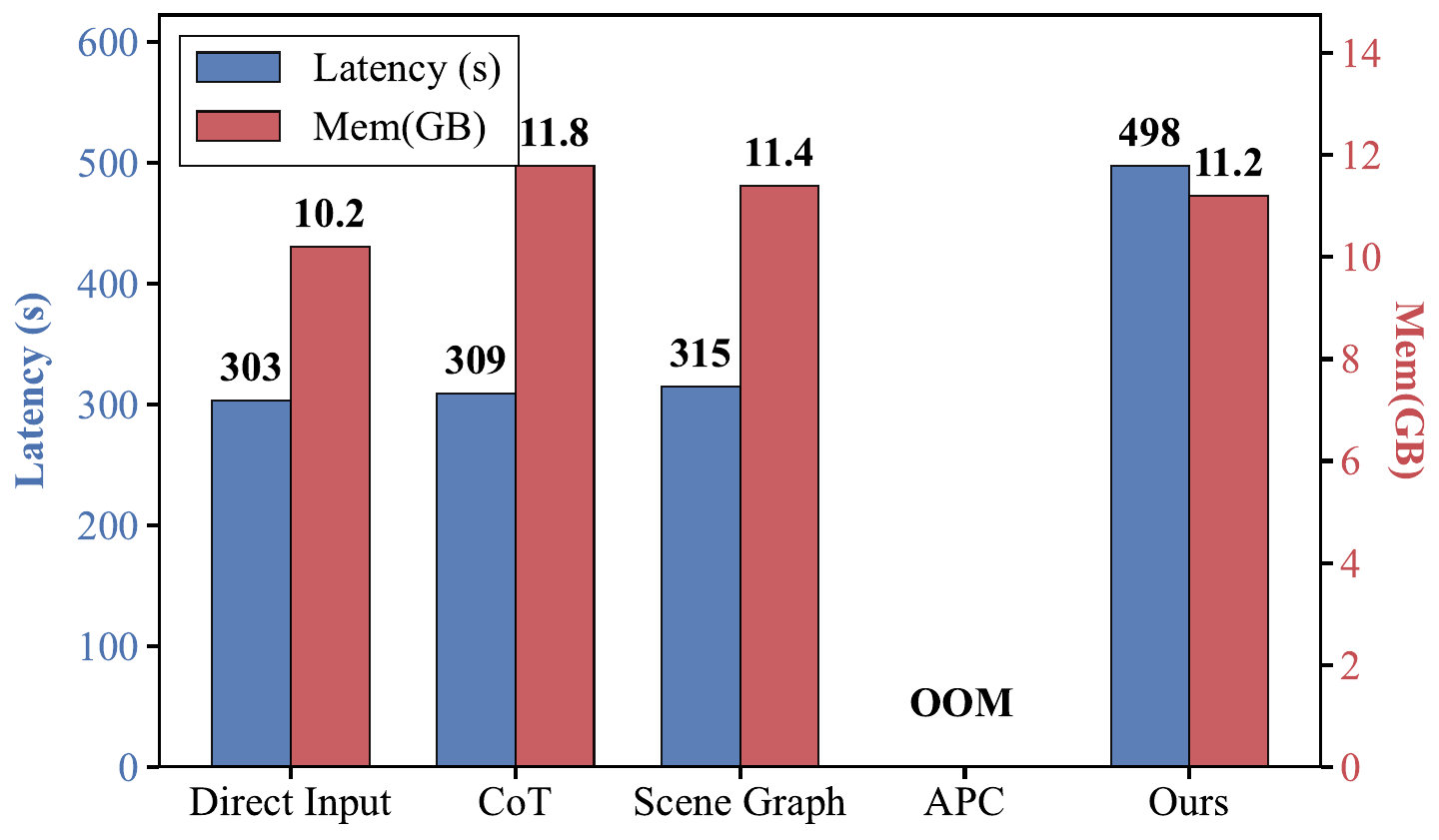}
        \label{fig:efficiency_3b}
    }
    \hspace{-0.25in}
    \hfill 
    \subfigure[Intern-VL3-8B on Jetson Orion]{
        \includegraphics[width=0.35\linewidth]{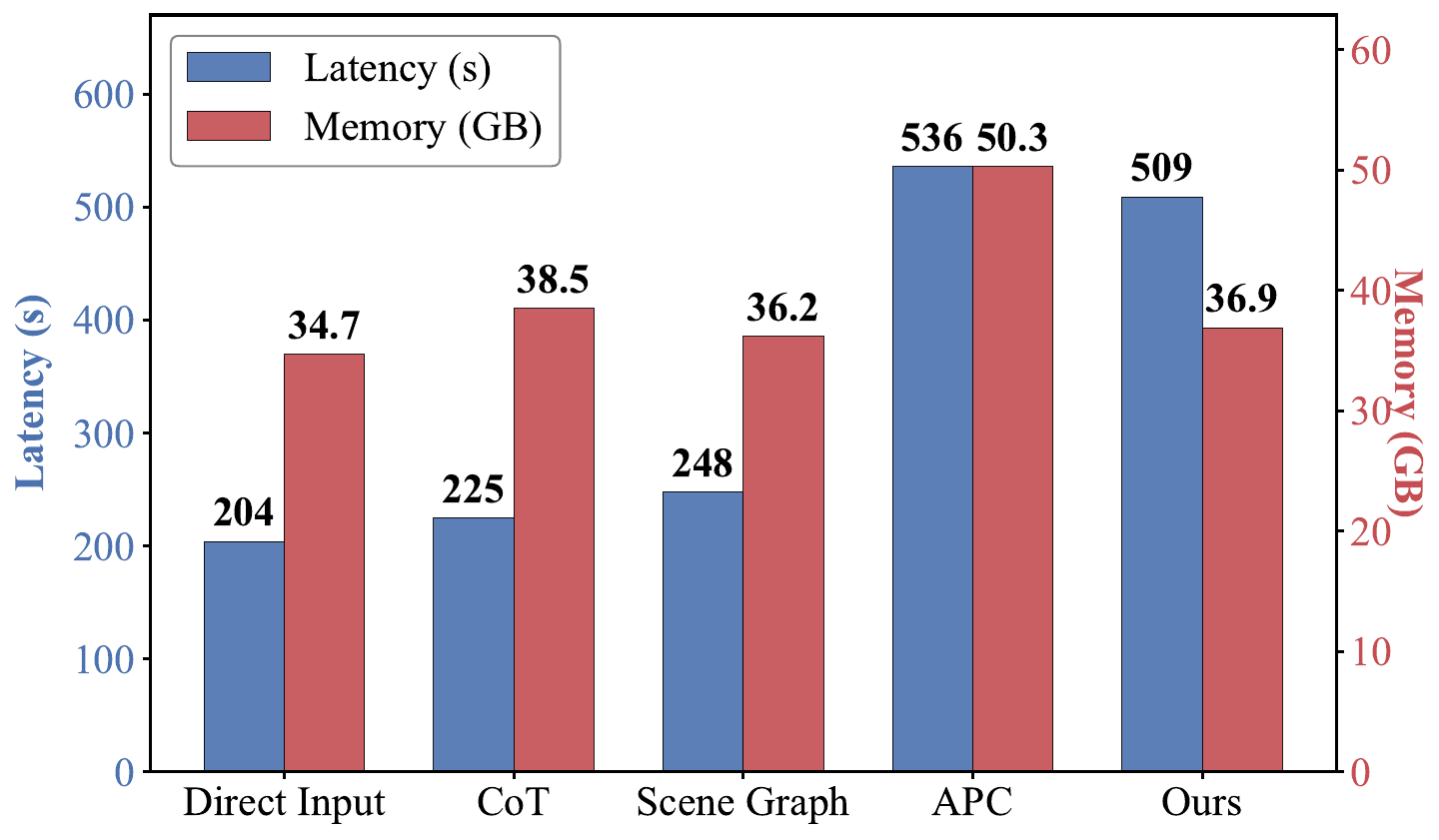}
        \label{fig:efficiency_8b}
    }
    \hspace{-0.25in}
    \hfill 
    \subfigure[Qwen3VL-32B-4bit on Jetson Orion]{
        \includegraphics[width=0.35\linewidth]{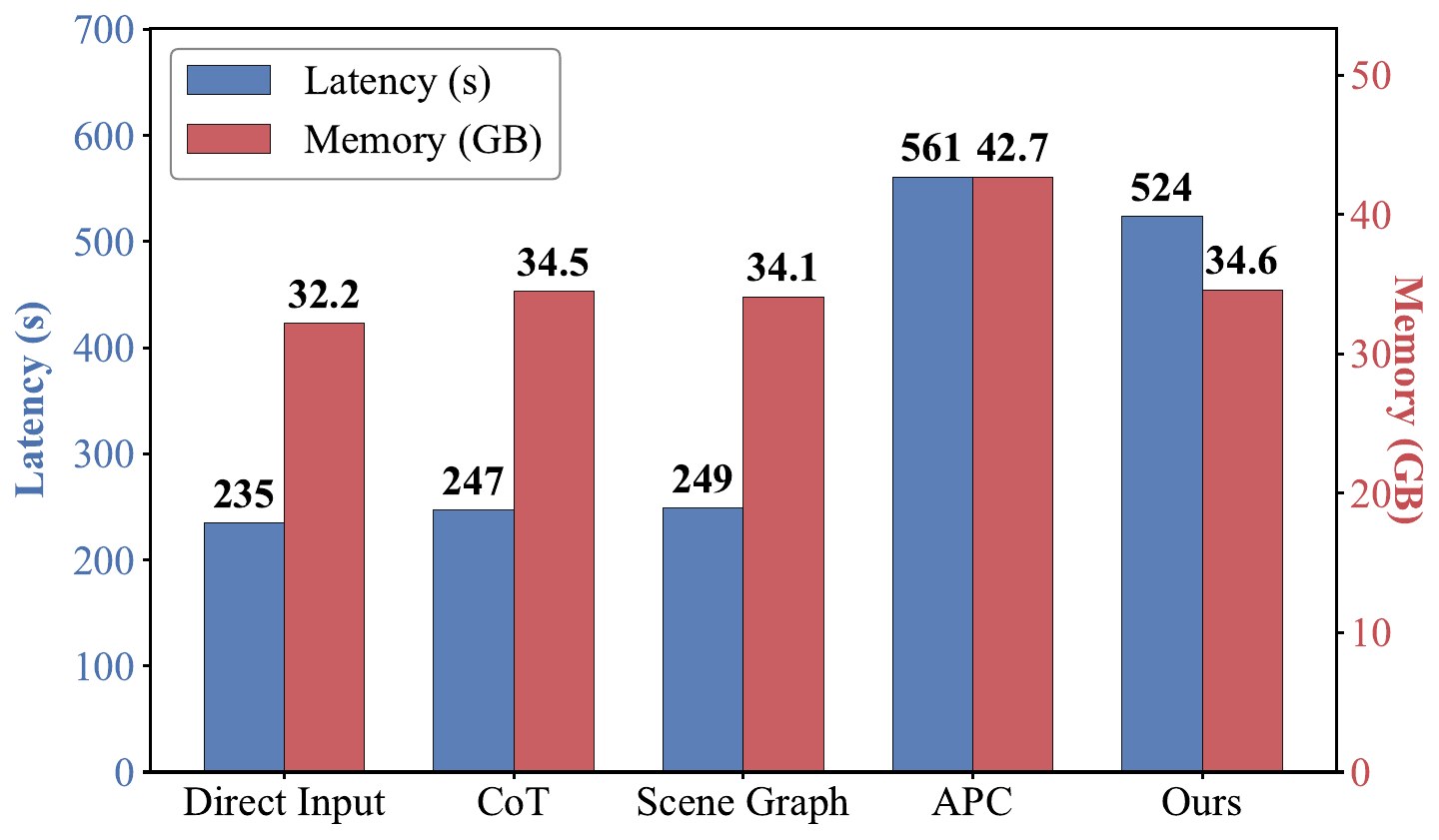}
        \label{fig:efficiency_32b}
    }
    \hspace{-0.25in}
    \vspace{-0.15in}
    \caption{Compute efficiency on different devices}
    \vspace{-0.0in}
    \label{fig:efficiency}
\end{figure}

In addition to the reasoning accuracy, Figure \ref{fig:efficiency} presents the compute latency and memory cost of different spatial reasoning methods when being deployed on the
NVidia Jetson Orion and Smartphones, which are critical considerations for deployment in embodied AI settings. 

As shown in Figure \ref{fig:efficiency}, MosaicThinker demonstrates robust feasibility on resource-constrained devices where the most competitive baseline (i.e., APC) fail. On the Smartphone (Fig. \ref{fig:efficiency_3b}), the APC method runs out of memory due to its inefficient incorporation of spatial information, whereas our approach successfully executes with 11.2GB memory usage, comparable to lighter baselines like Video-CoT. On the NVidia Jetson platform (Figs. \ref{fig:efficiency_8b} and \ref{fig:efficiency_32b}), MosaicThinker also significantly outperforms APC in resource efficiency, reducing memory consumption by over 26\% (from 50.3GB to 36.9GB on the 8B model) while also achieving faster inference (509s vs. 536s). While our method incurs higher latency than other baselines,  these inferior baselines trade the reasoning accuracy for speed, whereas MosaicThinker achieves much higher reasoning accuracy with a moderate amount of extra compute cost.

\subsection{Performance over Difference Levels of Scene complexities}

To evaluate the robustness of our method against in different environmental conditions, we analyze how the scene complexity affects the VLM's spatial reasoning performance on InternVL3-8B. We define the scene complexity along three aspects: (1) \textbf{object number} measured by the number of objects in the scene, (2) \textbf{spatial scale} measured by the room size, and (3) \textbf{temporal duration} measured by the length of video clip. For each factor, we stratify the test cases into three distinct levels: Low, Medium, and High.

We conduct these experiments using two complementary data sources to ensure both realism and precise controllability. While real-world datasets provide the necessary physical validity, they often suffer from inherent correlations among different complexity factors. For instance, larger physical spaces in real videos typically contain higher object counts and require longer durations to traverse. This entanglement makes it difficult to isolate the specific cause of performance degradation. Therefore, we also utilize synthetic data to decouple these variables.

\begin{wrapfigure}{r}{2.5in}
	\centering
		\vspace{-0.2in}
	\includegraphics[width=\linewidth]{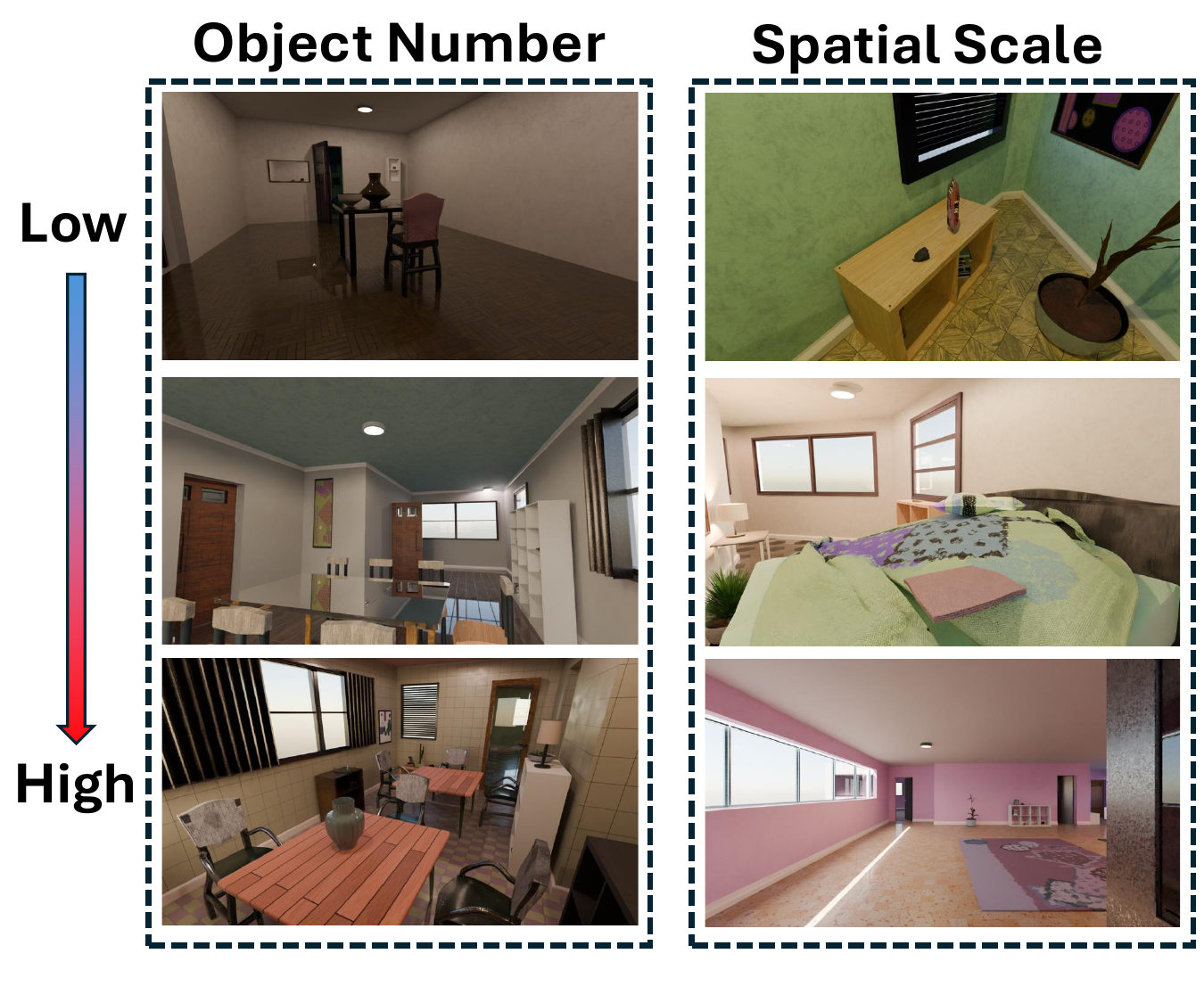}
	\vspace{-0.3in}
	\caption{Example of synthetic data samples with controlled complexity.}
	\label{fig:synthetic_eg}
\vspace{-0.05in}
\end{wrapfigure}

\begin{itemize}
    \item \textbf{Real-world data:} We utilize the VSI-Bench dataset \cite{yang2025thinking} which provides segmentation results of indoor scenes. We analyze the meta data to calculate scene statistics and categorize the existing test samples into the complexity levels specified above.
    
    \item \textbf{Synthetic Data:} To rigorously control for confounding variables, we also used a synthetic generation pipeline based on InfiniBench \cite{wang2025infinibench}. As shown in Figure \ref{fig:synthetic_eg}, this allows us to parametrically generate scenes with specific complexity configurations, render corresponding egocentric videos, and automatically generate valid spatial reasoning questions using their pipeline.
\end{itemize}

We compared our approach against two baselines: the standard \textit{Direct Input} method and \textit{APC}, which was the most competitive baseline identified in the previous subsection. For each experimental configuration, we report the average accuracy achieved by each method.

\begin{figure}[ht]
    \vspace{-0.1in}
    \centering
    \subfigure[Object Number]{
        \includegraphics[width=0.30\linewidth]{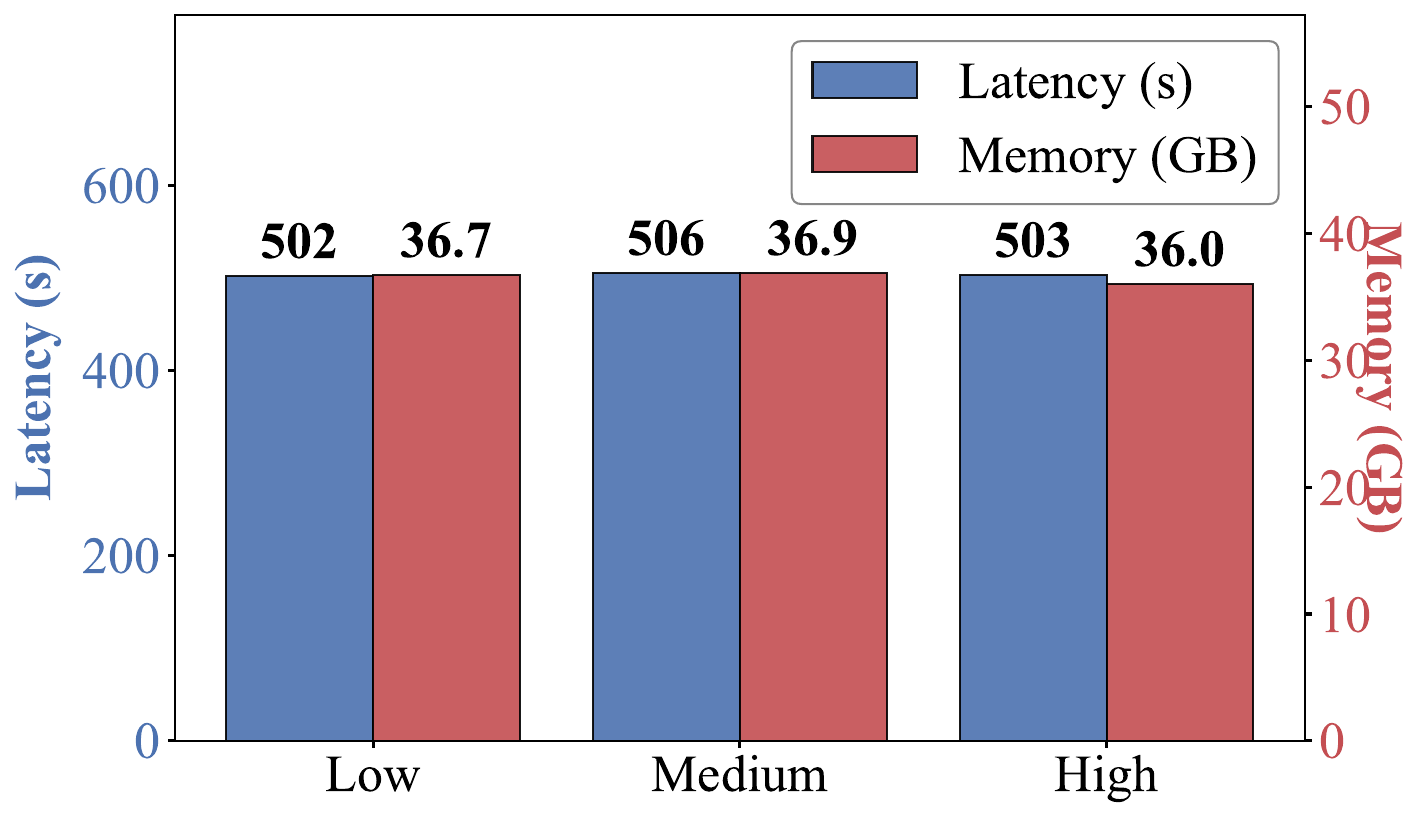}
        \label{fig:efficiency_number_3b}
    }
    \hspace{-0.25in}
    \hfill 
    \subfigure[Spatial Scale]{
        \includegraphics[width=0.30\linewidth]{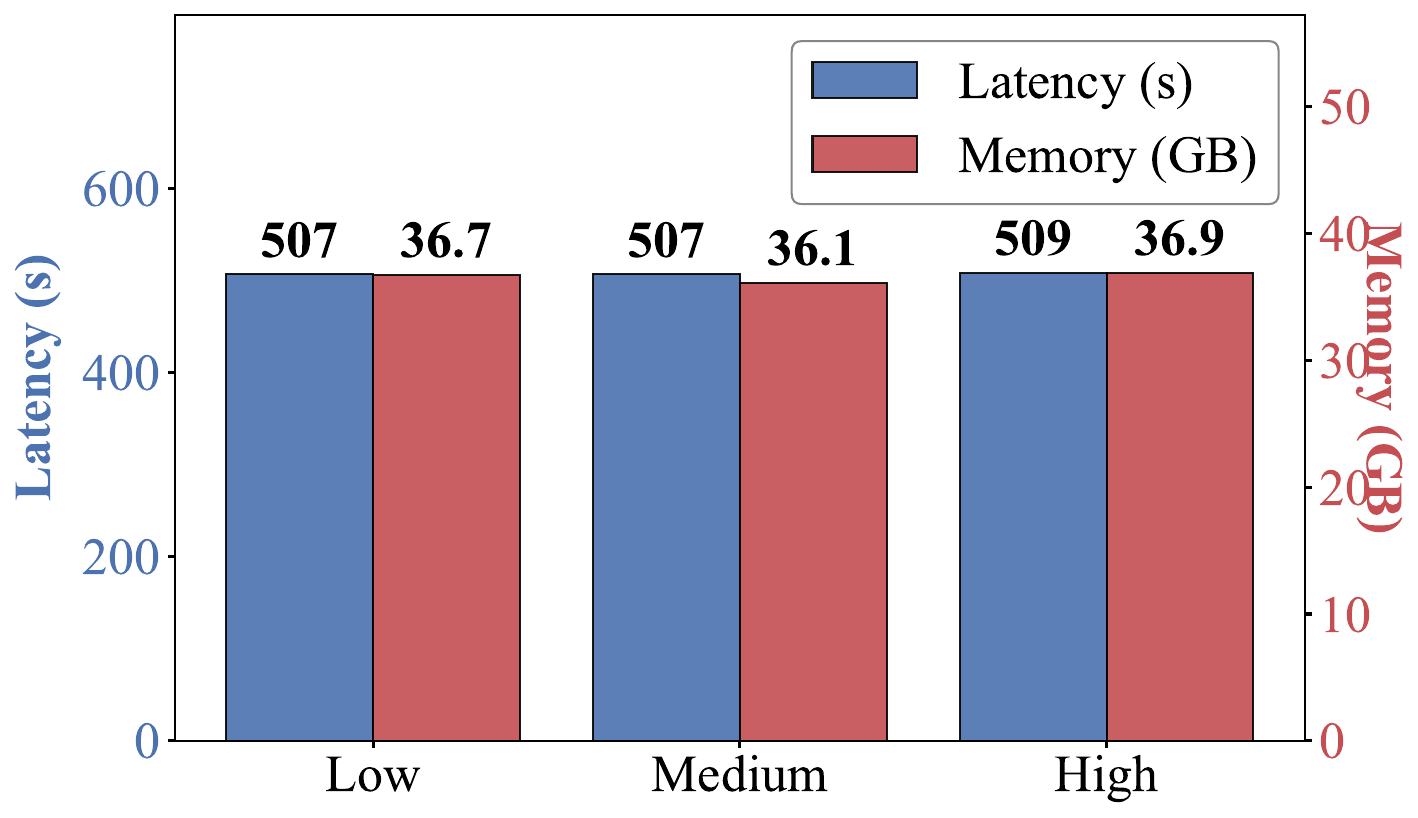}
        \label{fig:efficiency_scale_8b}
    }
    \hspace{-0.25in}
    \hfill 
    \subfigure[Temporal Duration]{
        \includegraphics[width=0.30\linewidth]{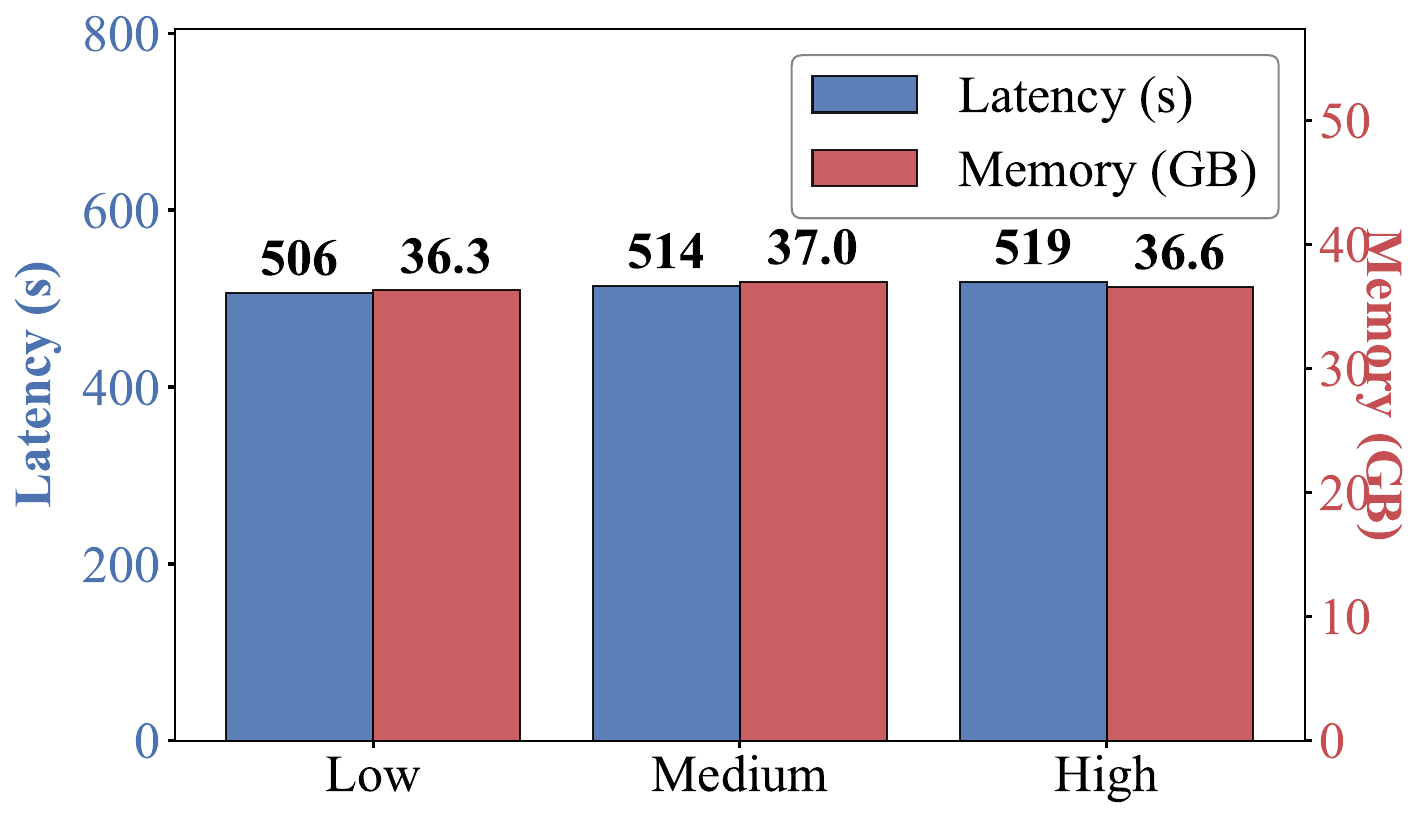}
        \label{fig:efficiency_duration_32b}
    }
    \vspace{-0.15in}
    \caption{Compute efficiency of MosaicThinker under different complexities}
    \vspace{-0.05in}
    \label{fig:efficiency_comlexity}
\end{figure}

Tables \ref{tab:complexity_real} and \ref{tab:complexity_synthetic} reveal that Object Number is the primary bottleneck; as the spatial reasoning accuracy drops sharply for all methods in cluttered scenes. However, MosaicThinker still performs better than the baselines. Crucially, MosaicThinker shows generalizability to Spatial Scale, outperforming baselines in large environments where all the baselines struggle. This confirms that our semantic map effectively integrates fragmented views, preserving spatial relationships in large space that are otherwise lost. Temporal Duration had the minimal impact, indicating that the key frame selection method is effective to retrieve necessary frames for spatial reasoning no matter the video length. Moreover, as shown in Figure \ref{fig:efficiency_comlexity}, the computational cost of MosaicThinker remains stable across different complexity levels; increased complexity does not introduce additional overhead.

\begin{table*}[ht]
\centering
\small
{%
\begin{tabular}{l ccc ccc ccc}
\toprule
& \multicolumn{3}{c}{\textbf{Object Number}} & \multicolumn{3}{c}{\textbf{Spatial Scale}} & \multicolumn{3}{c}{\textbf{Temporal Duration}} \\ 
\cmidrule(lr){2-4} \cmidrule(lr){5-7} \cmidrule(lr){8-10}
\textbf{Method} & Low & Med. & High & Low & Med. & High & Low & Med. & High \\ 
\midrule
Direct Input           & 53.2 & 47.8 & 32.5 & 50.3 & 48.9 & 35.1 & 33.1 & 35.7 & 32.6 \\
APC           & 54.6 & 47.7 & 32.2 & 52.1 & 47.2 & 34.3 & 33.5 & 35.3 & 31.8 \\
\textbf{MosaicThinker} & 60.2 & 53.9 & 39.2 & 62.5 & 54.8 & 41.6 & 41.2 & 39.9 & 38.4 \\
\bottomrule
\end{tabular}
}
\caption{Experiment results of VSI-Bench on different levels of scene complexity }
\vspace{-0.2in}
\label{tab:complexity_real}
\end{table*}

\begin{table*}[h]
\centering
\small
{%
\begin{tabular}{l ccc ccc ccc}

\toprule
& \multicolumn{3}{c}{\textbf{Object Number}} & \multicolumn{3}{c}{\textbf{Spatial Scale}} & \multicolumn{3}{c}{\textbf{Temporal Duration}} \\ 
\cmidrule(lr){2-4} \cmidrule(lr){5-7} \cmidrule(lr){8-10}
\textbf{Method} & Low & Med. & High & Low & Med. & High & Low & Med. & High \\ 
\midrule
Direct Input           & 67.9 & 44.8 & 37.1 & 52.0 & 51.1 & 46.8 & 63.8 & 66.9 & 65.3 \\
APC           & 69.8 & 43.9 & 36.0 & 52.9 & 51.2 & 48.0 & 62.5 & 67.0 & 65.5 \\
\textbf{MosaicThinker} & 79.7 & 54.2 & 42.7 & 61.5 & 58.1 & 55.2 & 71.2 & 70.5 & 69.3 \\
\bottomrule
\end{tabular}
}
\caption{Experiment results of InfiniBench on different levels of scene complexity }
\vspace{-0.25in}
\label{tab:complexity_synthetic}
\end{table*}

\subsection{Ablation Studies}

\textbf{Impact of Keyframe Selection Strategy.} To validate the effectiveness of our iterative temporal search, we compare it against two baselines: (1) processing all frames (at 1 FPS) and (2) uniform sampling, on Jetson Orion with Inverl-VL3-8B and Qwen3VL-32B-4bit using VSI-bench. As shown in Table \ref{tab:keyframe_ablation}, processing all frames results in Out-of-Memory; even in simulation on high-memory hardware, accuracy degrades due to the excessively long context window. Conversely, while uniform sampling is computationally feasible, it suffers from lower accuracy by missing critical, task-relevant frames. Our iterative approach achieve the optimal balance between accuracy and computational cost by concentrating resources on semantically significant frames.
\begin{table}[h]
\centering
\small
\begin{minipage}[t]{0.48\textwidth}
    \centering
    \begin{tabular}{lcc}
    \toprule
    \textbf{Intern-VL3-8B} & \textbf{Accuracy} & \textbf{Latency} \\ \midrule
    All frames (1 FPS)        & 43.2\%            & OOM              \\
    Uniform sampling          & 46.0\%            & 389 s            \\
    \textbf{Ours (Iterative)} & \textbf{58.7\%}   & 501 s            \\ \bottomrule
    \end{tabular}
    \label{tab:ablation_selection_8B}
\end{minipage}
\hfill 
\begin{minipage}[t]{0.48\textwidth}
    \centering
    \begin{tabular}{lcc}
    \toprule
    \textbf{Qwen3VL-32B-4bit} & \textbf{Accuracy} & \textbf{Latency} \\ \midrule
    All frames (1 FPS)        & 59.5\%            & OOM              \\
    Uniform sampling          & 62.0\%            & 409 s            \\
    \textbf{Ours (Iterative)} & \textbf{73.3\%}   & 524 s            \\ \bottomrule
    \end{tabular}
    \label{tab:ablation_selection_32B}
\end{minipage}
\caption{Ablation studies on different keyframe selection strategies.}
\vspace{-0.05in}
\label{tab:keyframe_ablation}
\end{table}

\noindent\textbf{Efficacy of Visual Prompting Methods.} We further investigate the impact of different prompting methods on VLM reasoning performance. Table \ref{tab:representation_formats} compares the prompting methods used in MosaicThinker (Semantic map + Text) against two alternatives: projecting dense point clouds directly as visual inputs and using pure text-based coordinate descriptions. The experiments were conducted using InternVL3-8B and Qwen3VL-32B-4bit on the Jetson Orin with the VSI-Bench dataset. The results indicate that the semantic map is the most effective input format. It significantly outperforms text descriptions which VLMs struggle to interpret spatially, and dense point clouds which can introduce visual noise. Our method enhances reasoning accuracy without imposing a substantial latency overhead compared to simpler baselines.

\begin{table}[h]
\centering
\small
\begin{minipage}[t]{0.48\textwidth}
    \centering
    \begin{tabular}{lcc}
\toprule
\textbf{Intern-VL3-8B} & \textbf{Accuracy} & \textbf{Latency} \\ \midrule
Dense Points (BEV)           & 46.1\%            & 172 s           \\
Text Description             & 45.2\%            & 169 s           \\
\textbf{Semantic map + text} & \textbf{58.7\%}   & 172 s      \\ \bottomrule
\end{tabular}
    \label{tab:ablation_prompt_8B}
\end{minipage}
\hfill 
\begin{minipage}[t]{0.48\textwidth}
    \centering
    \begin{tabular}{lcc}
\toprule
\textbf{Qwen3VL-32B-4bit} & \textbf{Accuracy} & \textbf{Latency} \\ \midrule
Dense Points (BEV)           & 61.3\%            & 202 s           \\
Text Description             & 61.0\%            & 187 s           \\
\textbf{Semantic map + text} & \textbf{73.3\%}   & 201 s      \\ \bottomrule
\end{tabular}
    \label{tab:ablation_prompt_32B}
\end{minipage}
\caption{Ablation studies on different formats for VLM prompting.}
\label{tab:representation_formats}
\end{table}


	\vspace{-0.05in}
	\section{Related Work}
	
	\textbf{Embodied Spatial Intelligence}. Besides language intelligence that has been well supported by LLMs, spatial intelligence has recently emerged as a research frontier at the intersection of computer vision, AI, robotics and mobile systems \cite{yamada2023evaluating,sharma2023exploring,yang2025thinking,bhandari2023large}. Unlike traditional AI systems that operate on abstract symbolic inputs without physical grounding, embodied spatial intelligence focuses on spatial perception and understanding within the 3D world. It enables agents to acquire and process spatial information to construct internal representations, and to use these representations to guide navigation, task execution, and decision-making \cite{feng2025survey}. Recent advances have driven progress in applications of robotic navigation, manipulation, AR/VR, and autonomous driving \cite{lin2023advances,feng2025survey}. However, despite these achievements, existing methods still struggle with constructing robust spatial representations \cite{yang2025thinking}. To address this, in this paper, we propose MosaicThinker, a framework that reconstructs a semantic map to serve as an effective spatial representation.
	
	\noindent \textbf{Training-based approaches to spatial reasoning}. In this paper, we focus on training-free approaches to improving capabilities of spatial reasoning. If users are willing to incur additional computational cost and accept a potential loss in generalizability, they may fine-tune the model on spatial data enriched with additional modalities, such as depth maps, or employ architectures specifically designed for spatial reasoning \cite{zheng2025video,yu2025inst3d,wu2025spatial,chen2024ll3da,yang2025lidar}. Notably, in MosaicThinker, the generated semantic map can itself serve as supervision for model fine-tuning, further enhancing performance by teaching the model to construct and interpret semantic maps \cite{qi2025gpt4scene,yin2025spatial}.
	
	\vspace{-0.05in}
	\section{Discussions}
	
	\textbf{Spatial reasoning tasks in other forms.} With the given visual inputs, in this paper we target spatial reasoning tasks with questions in natural language, but these tasks can also be specified in other ways. For example, the tasks are not necessarily QA, but could also be descriptions of required actions or sequences of actions. Similarly, the VLM's outputs, as the outcomes of task completion, are not necessarily answers in text, but could also be multimodal such as bounding boxes, image patches, or robotic actuations \cite{yang2022unitab,zitkovich2023rt}. MosaicThinker can also be applied to these tasks, by adapting its output decoder to generate structured data directly, or mapping its internal spatial representations to specific formats like coordinates, image patches, or action primitives.
	
	\vspace{0.05in}
	\noindent\textbf{Spatial reasoning in large-scale environments.} While the experiments in this paper focus on spatial reasoning in small-scale indoor environments, MosaicThinker can scale up by design. In large-scale environments, such as warehouses and hospitals, MosaicThinker can continuously build a comprehensive spatial memory as the robot explores. By leveraging techniques such as Retrieval-Augmented Generation (RAG) to retrieve question-relevant portions of this memory \cite{xie2024embodied,booker2024embodiedrag}, we can answer complex spatial queries, navigate efficiently, and adapt its behavior based on accumulated knowledge of the environment.
	
		\vspace{-0.05in}


	\section{Conclusion}
	In this paper, we focus on enhancing the ability of small on-device VLMs on cross-frame visual spatial reasoning tasks, which are the key enabler of emerging embodied AI applications such as robotic manipulation and actuation planning. We present a new inference-time computing technique, namely MosaicThinker, which integrates the fragmented spatial information from multiple video frames into a unified space representation of global semantic map. This semantic map is used as VLM input for spatial reasoning, guided by a carefully crafted visual prompt. Experiment results on multiple embodied AI devices and spatial reasoning benchmarks show that MosaicThinker largely outperforms the competitive baselines, in various types of indoor scenes and spatial reasoning tasks.
\setcitestyle{numbers}
\bibliographystyle{abbrvnat}
\bibliography{main}

\newpage
\appendix

\section{Representing the Semantic Map to VLM}
\label{subsec:visual_prompt_details}

\begin{wrapfigure}{r}{3.5in}
	\centering
	\includegraphics[width=\linewidth]{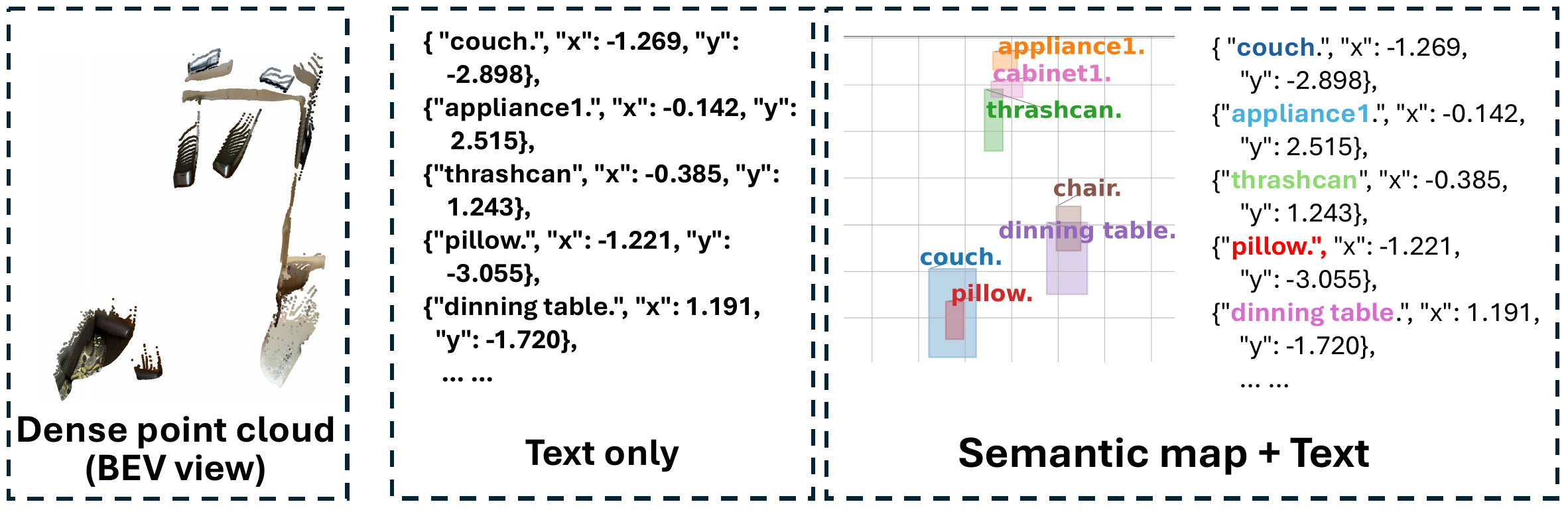}
	\caption{Different formats of space representation as inputs to VLMs}
	\label{fig:different_inputs}
	\vspace{-0.15in}
\end{wrapfigure}
With the semantic map, we still need to determine the most effective representation of the semantic map as the input to VLM. Representing the semantic map directly in structured text (e.g., JSON) has been proved to perform poorly on small VLMs \cite{yin2025spatial}, since the VLM has to internally translate numerical values into a visual representation, which is a shortcoming of most language models nowadays. For example, a statement such as ``Object A is 2 meters to the north of Object B, facing east'' is immediately apparent in a map, but would be verbose and less intuitive in JSON. This difficulty motivates us to instead adopt visual prompting techniques \cite{lei2024scaffolding,liu2024coarse} by plotting the semantic map as an image, as illustrated in Figure \ref{fig:different_inputs}.

\section{Temporal Locality among Video Frames}
\label{appendix:temporal_locality}

\begin{wrapfigure}{r}{3in}
	\centering
	\vspace{-0.2in}
	\includegraphics[width=\linewidth]{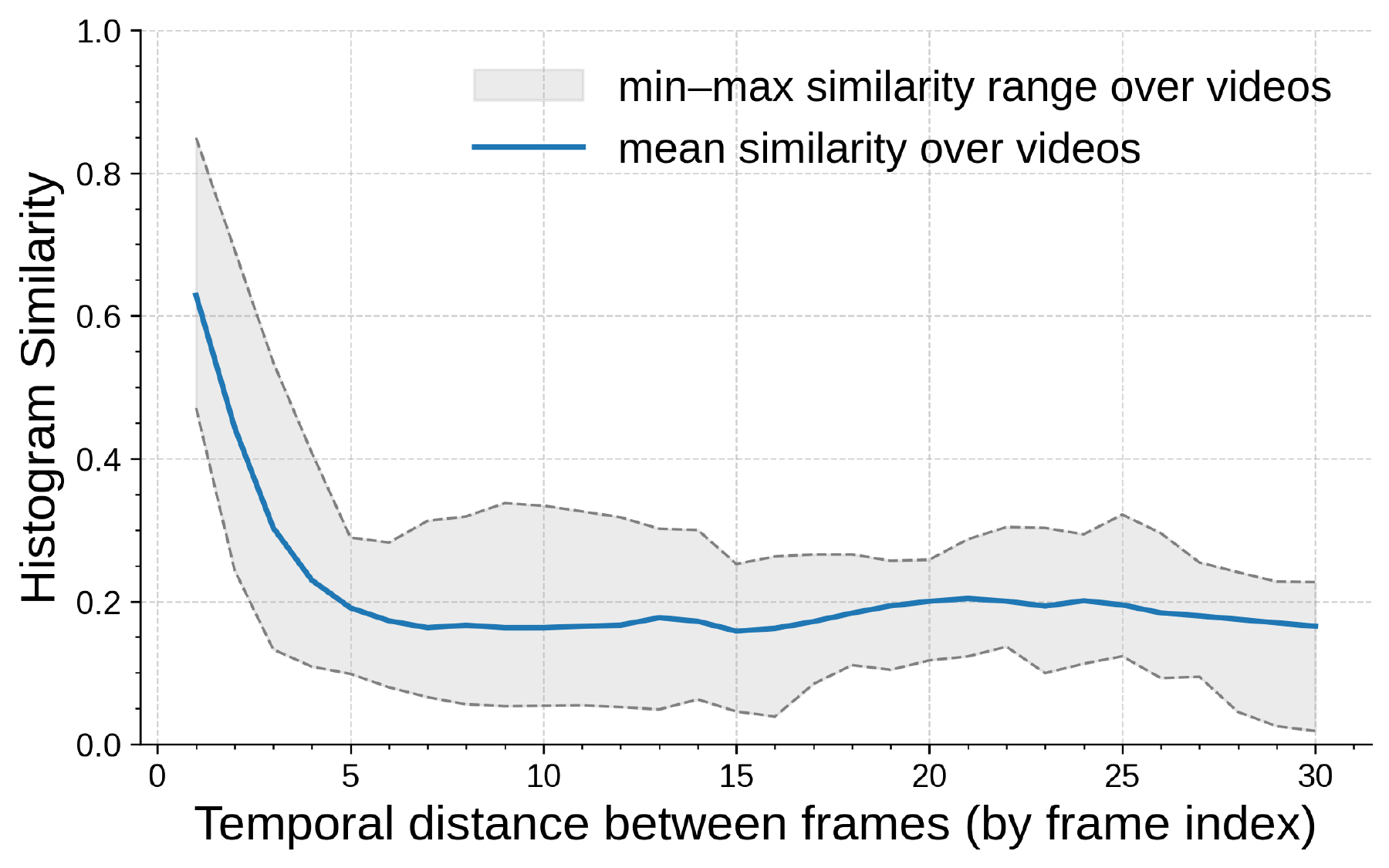}
	\vspace{-0.25in}
	\caption{Histogram frame similarity with different temporal distances between frames}
	\label{fig:temporal_locality}
	\vspace{-0.1in}
\end{wrapfigure}
Our proposed approach of iterative temporal search for key frame selection in Section \ref{subsec:keyframe_selection} builds on the temporal locality of object appearance among video frames. That is, if a frame is highly likely to contain an object, other frames that are temporally close could also have high likelihoods to contain the same object, due to the continuity of camera movement in the video. 

To experimentally validate such locality, we randomly sampled 100 videos from the VSIBench benchmark that depict indoor scenes \cite{yang2025thinking}, randomly selected 10\% frames from each video as reference frames, and computed the average histogram similarity \cite{zweng2011evaluation} between each reference frame and its subsequent frames at different temporal distances. As shown in Figure~\ref{fig:temporal_locality}, the frame similarity remains high in the first 2-3 frames, and then rapidly decays to a low level. This result suggests strong temporal locality within the range of 2-3 frames, which supports our approach of iterative temporal search, and is also used in Section  \ref{subsec:gaussian_kernel} to decide the width of the Gaussian kernel to be applied when updating the sampling distribution over iterations.

\section{Details of the Metro-Spatial-QA Dataset}
\label{appendix:dataset}
\begin{figure*}
    \includegraphics[width=0.85\textwidth]{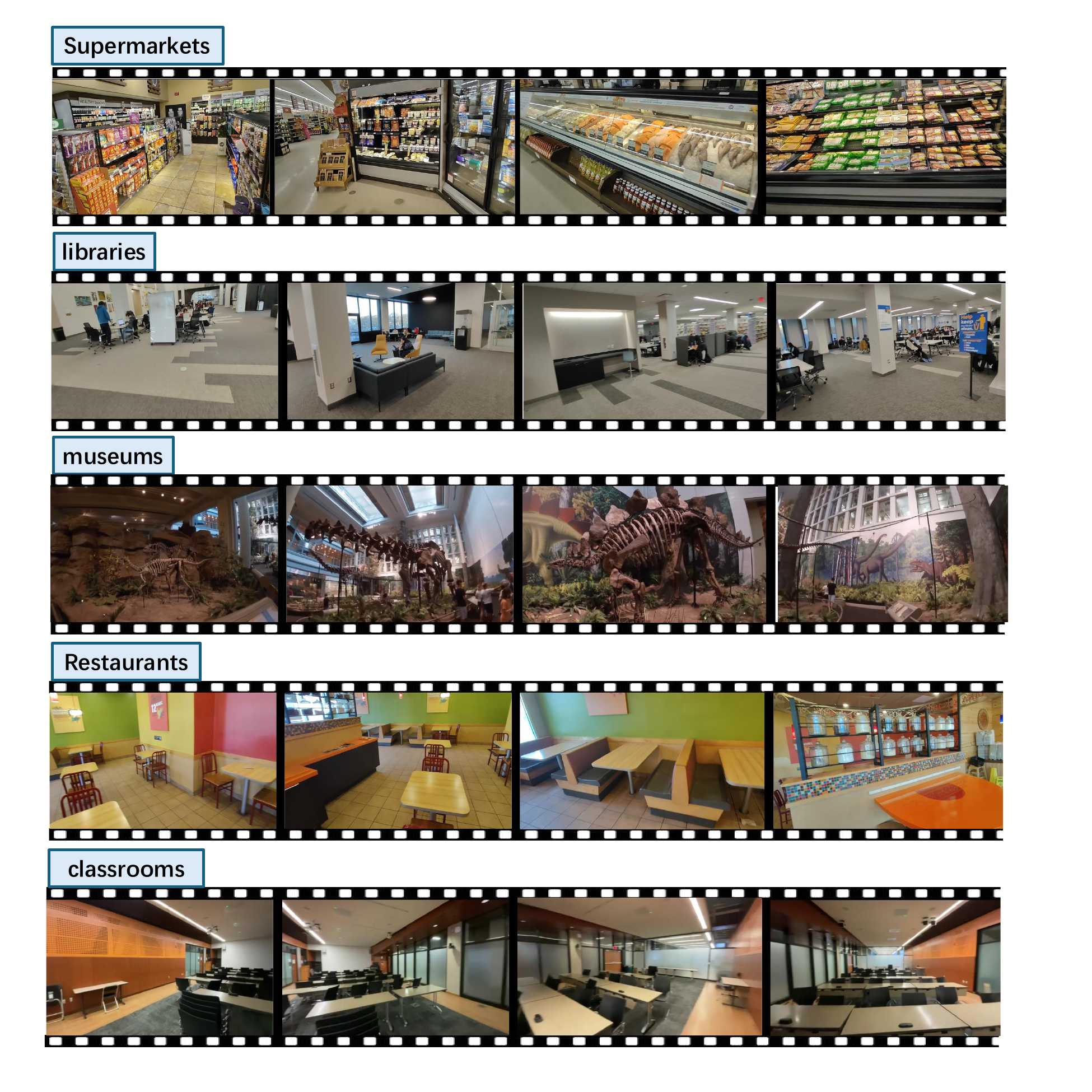}
    \vspace{-0.1in}
    \caption{Data samples from Metro-Spatial-QA}
    \label{figself_collected}
\end{figure*}

We collect a new egocentric video dataset containing 40 egocentric video clips, covering a diverse set of indoor environments commonly encountered in daily activities. As shown in Figure \ref{figself_collected}, the dataset spans five categories of places: supermarkets, libraries, museums, restaurants and classrooms. These environments exhibit substantial variation in spatial layout, object distribution, lighting conditions, and levels of human activity, providing rich visual and contextual cues for spatial understanding from egocentric video streams. For each clip, we manually annotated 4 spatial reasoning questions, yielding 160 spatial reasoning tasks in total.
The collected scenes are intentionally diverse in both structure and complexity, ranging from highly organized spaces such as restaurants to cluttered and dynamic environments such as supermarkets and museums. In addition, classroom and laboratory settings introduce semi-structured layouts with functional objects and equipment arranged for specific tasks. This diversity leads to challenging spatial configurations involving occlusions, long-range relationships, and viewpoint changes across frames. Consequently, the dataset provides a realistic and challenging testbed for evaluating visual spatial reasoning and cross-frame understanding in real-world indoor environments.

\end{document}